\documentclass[5p, times]{elsarticle}

\usepackage{ecrc}
\usepackage[inline]{enumitem}
\usepackage{comment}
\usepackage{graphicx} 
\usepackage{amsmath} 
\usepackage{amsfonts}
\usepackage{mathtools}
\usepackage{microtype}
\usepackage{subfig}
\usepackage{tikz}
\usepackage{adjustbox}
\usepackage{booktabs}
\usepackage{multirow}
\usepackage{url}
\usepackage{svg}
\usepackage{optidef}
\usepackage{xcolor,colortbl}

\newcommand{\metric}[1]{\textit{#1}}

\usepackage{caption}

\usepackage{booktabs}
\usepackage{array}
\usepackage{tabularx}
\usepackage{makecell}
\newcolumntype{Y}{>{\raggedright\arraybackslash}X}
\newcolumntype{L}[1]{>{\raggedright\arraybackslash}p{#1}}

\usepackage{algorithm}

\usepackage{algpseudocode}
\usepackage{hyperref}
\hypersetup{
 colorlinks=true,
 linkcolor=blue,
 filecolor=magenta, 
 urlcolor=cyan,
 citecolor= gray,
 pdftitle={Overleaf Example},
 pdfpagemode=FullScreen,
 pdfauthor=whatever
 }

\usepackage[separate-uncertainty=true]{siunitx}
\usepackage{dirtytalk}

\usepackage{threeparttable} 

\usepackage{colortbl}
\usepackage{multirow}
\usepackage{makecell}

\usepackage{romannum}

\usepackage{flushend}
\usepackage{tabularx}

\usepackage{soul} 

\usepackage[normalem]{ulem} 

\volume{0}

\firstpage{0}

\journalname{Robotics and Computer-Integrated Manufacturing}

\runauth{et al.}

\usepackage{amssymb}

\definecolor{Gray}{gray}{0.95}
\definecolor{LightGray}{gray}{0.95}

\definecolor{LightCyan}{rgb}{0.88,1,1}

\definecolor{TableHeader}{gray}{0.94}
\definecolor{TableSection}{gray}{0.97}
\definecolor{TableAccent}{RGB}{232,241,247}
\newcommand{\tableheaderrow}{\rowcolor{TableHeader}}
\newcommand{\tablesectionrow}{\rowcolor{TableSection}}
\newcommand{\tableaccentrow}{\rowcolor{TableAccent}}

\newcolumntype{G}{>{\columncolor{Gray}}c}
\begin{document}

\begin{frontmatter}

 \title{MaCoPlanner: LLM-Assisted Manual-Compiled Task Planning with Proactive Safety Verification for Robotic Industrial Panel Operation}

\author[HUST]{Guipeng Xin\fnref{fn1}}
\ead{xinguipeng@hust.edu.cn}

\author[HUST]{Jiahe Xu\fnref{fn1}}
\ead{jiahexu@hust.edu.cn}

\author[UNSW]{Mohammad Deghat}
\ead{M.deghat@unsw.edu.cn}

\author[HUST]{Chenhui Wan}
\ead{wanchenhui@hust.edu.cn}

\author[HUST1]{Jie Liu}
\ead{jie\_liu@hust.edu.cn}

\author[HUST]{Youmin Hu}
\ead{youmhwh@hust.edu.cn}

\author[HUST]{Zhongxu Hu\corref{cor1}}
\ead{zhongxu\_hu@hust.edu.cn}

\cortext[cor1]{Corresponding Author.}
\fntext[fn1]{These authors contributed equally to this work.}

\address[HUST]{School of Mechanical Science and Engineering, Huazhong University of Science and Technology, Wuhan, China}
\address[UNSW]{School of Mechanical and Manufacturing Engineering, University of New South Wales, Sydney, Australia}
\address[HUST1]{School of Civil and Hydraulic Engineering, Huazhong University of Science and Technology, Wuhan, China}

\begin{abstract}
Robotic industrial panel operation requires not only accurate control localization but also compliance with operating procedures, safety rules, and device-state constraints distributed across heterogeneous manuals. This study presents \emph{MaCoPlanner}, a task-planning framework built on knowledge compiled from equipment manuals that converts equipment manuals into a typed intermediate representation, retrieves task- and state-relevant evidence, and uses it to support plan generation. Before actuation, candidate plans are symbolically rolled out and checked against procedural and state-transition constraints; detected violations are localized and returned for targeted repair, while unresolved plans are rejected. A separate execution interface grounds verified symbolic actions to physical controls and updates the device state. Under an independent evaluation oracle, MaCoPlanner achieves a final violation rate of 2.7\%, and 26.3\% of the runs in the repair analysis are rejected after exhausting the refinement budget. Compared with Raw-Manual, task success increases from 62.8\% to 84.4\% on Level-2 tasks and from 25.9\% to 43.2\% on Level-3 tasks. Experiments on a controller-panel simulator without an attached industrial load further demonstrate integrated execution feasibility under representative interaction conditions, without claiming industrial deployment readiness.
\end{abstract}

\end{frontmatter}

\section{Introduction}\label{sec: intro}

Industrial equipment operation panels are widely used in production sites as the primary human-machine interface (HMI) for device monitoring and control. Through buttons, switches, knobs, and display units, operators directly change equipment states and thus affect the stability and safety of industrial processes \cite{mourtzis2023future}. In high-risk or high-cost scenarios, such as live-line maintenance, fault handling and recovery in substations, and hazardous industrial environments, using robots to replace human workers in operating industrial panels is highly desirable. Yet current industrial panel automation still relies heavily on expert programming and manually designed procedures. These solutions usually involve high deployment cost and limited adaptability, since they are mainly designed for fixed devices and predefined workflows and are difficult to transfer under device variation, small-batch production, rapid changeover, and abnormal event handling \cite{pan2012recent}.

\begin{figure*}[t]
 \centering
 \makebox[\linewidth][c]{\includegraphics[trim={0cm 0cm 0cm 0cm},clip,width=0.75\textwidth]{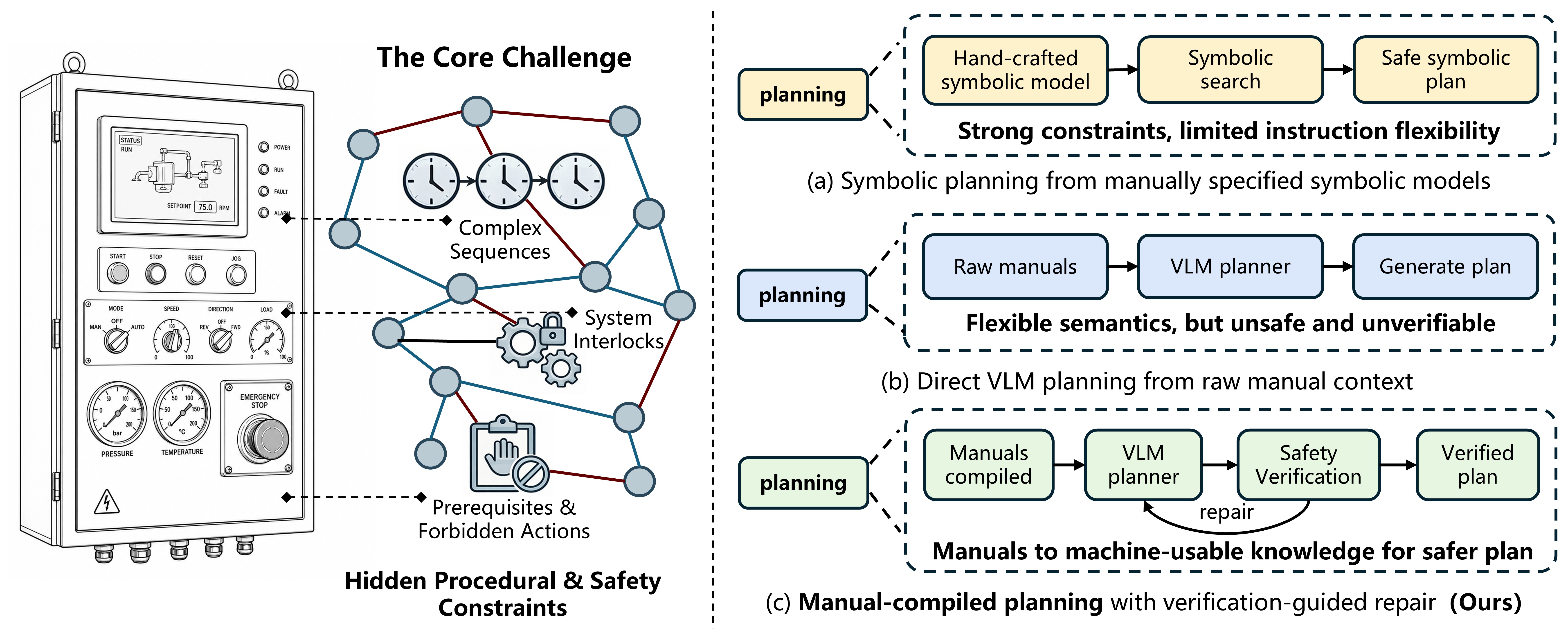}}
\caption{Core challenge and planning-paradigm comparison for robotic industrial panel operation. The left part highlights hidden procedural, state-dependent, and safety constraints. The right part contrasts classical symbolic planning, direct VLM planning, and the proposed planning paradigm based on compiled equipment-manual knowledge.}
\label{sys_overview}
\end{figure*}

Meanwhile, recent foundation-model-based robotic systems have shown strong capabilities in language understanding, multimodal perception, and long-horizon reasoning \cite{liang2022code, huang2023voxposer, yang2025guiding, hu2023look, mon2025embodied}. These advances open up the possibility of generating robotic task plans directly from unstructured user instructions \cite{fu2024can, li2023can}.
However, many of these systems are primarily developed and evaluated in household or general embodied manipulation settings, where task planning is relatively less dependent on device-specific operating logic. 
Compared with such tasks \cite{yin2024safeagentbench}, industrial panel operation is more strongly governed by operating procedures, device states, system interlocks, and safety requirements, while the corresponding knowledge is distributed across operation manuals, safety procedures, and parameter specifications \cite{ren2024embodied}. 
As illustrated in Figure~\ref{sys_overview}, directly feeding heterogeneous manuals into a VLM context does not reliably capture these distributed dependencies, particularly as the manual collection grows, and the generated plan may still contain procedural or state inconsistencies \cite{lou2024knowledge, fan2025embodied, ren2024embodied}. 
This motivates the central problem considered in this work: how to transform human-oriented equipment manuals into machine-usable task knowledge and rules that can directly support flexible robot planning and reduce unsafe plan generation.

To address this problem, this study presents \emph{MaCoPlanner}, a task-planning framework based on knowledge compiled from heterogeneous equipment manuals into a typed intermediate representation (IR) for robot task planning.
The compiled IR represents procedures, task-level safety constraints, state conditions, parameter specifications, and control relations in machine-usable forms, with source back-links retained for evidence retrieval, plan generation, symbolic rollout, verification, and repair.
In this way, equipment manuals are used as structured planning knowledge rather than only as auxiliary prompt context \cite{li2025vlm, kim2024fly, tie2025manual2skill, zhang2025robot}.
Based on the compiled representation, \emph{MaCoPlanner} performs proactive task-level verification before physical execution. Candidate plans are checked against procedural and state-dependent constraints, and detected violations are returned to the planner for targeted repair. 
\emph{MaCoPlanner} outputs a verified symbolic action sequence or rejects the task if verification fails within the refinement budget. This preserves flexible instruction understanding while improving task-level safety through pre-execution verification.

The key contributions of this study are summarized as follows:
\begin{enumerate}
\item \emph{MaCoPlanner} was proposed to compile heterogeneous manuals into structured machine-usable evidence, providing a more structured task-planning paradigm for knowledge-intensive industrial scenarios than direct long-context prompting with raw manuals, and supporting adaptation to new equipment through manual-derived task knowledge.

\item An evidence-conditioned proactive safety verification mechanism was developed to jointly enforce procedural constraints and mode/interlock constraints, thereby reducing unsafe plan generation before execution and rejecting action sequences that cannot satisfy the encoded task-level constraints within the refinement budget. 

\item A controlled evaluation chain spanning manual-knowledge compilation, retrieval quality, downstream task planning, proactive safety verification, and physical robot execution was established, and the proposed framework was validated on benchmark tasks and a representative robotic platform, thereby evaluating the quality and downstream effect of the manual-to-machine knowledge transformation and supporting integrated robotic panel-operation feasibility under the evaluated interaction conditions.
\end{enumerate}

\section{Related Works}

\subsection{Robotic Equipment Operation}
Research on robotic equipment operation mainly involves perception and manipulation. On the perception side, existing methods often rely on task-specific datasets and supervised learning for control instance segmentation, meter reading, and nameplate recognition \cite{liu2021large, fan2022real, verzic2024recovering}. These methods can be effective in constrained settings, but their performance is still limited by dataset scale, annotation quality, and scene coverage. As a result, their cross-device generalization remains insufficient in real industrial environments. On the manipulation side, conventional approaches usually build executable policies based on motion primitives, imitation learning, and reinforcement learning \cite{saveriano2023dynamic, zare2024survey, tang2025deep}. Recent advances in foundation models have also promoted more general manipulation methods, such as AnyGrasp \cite{fang2023anygrasp}, which improve object grasping across diverse targets. However, these methods mainly target object pickup and grasp pose generation, and are not directly suitable for fine-grained control operations such as button pressing, knob turning, and switch manipulation. Some recent studies further explore operation with large vision-language models (LVLMs) by inferring spatial constraints and predicting action locations for unseen tasks \cite{shao2025large}. Although this improves adaptability, general foundation models remain sensitive to spatial relations and physical properties \cite{gao2024physically}, and still fall short in precise and repeatable device operation. Robotic equipment operation has also been extended to long-horizon and multi-step tasks, such as domestic and industrial device manipulation \cite{yang2026instrucrobo, zhang2025robot}, chemical laboratory automation \cite{dai2024autonomous}, and flight simulator control \cite{kim2024fly}. However, many methods still depend on device-specific priors, one-time calibration, or fixed workflows, which limits cross-device and cross-task transfer.

\subsection{LVLM-Based Task Planning}
Applying LVLMs to robotic planning, especially high-level task planning, has significantly changed the planning paradigm. Classical symbolic planning is effective when a complete planning domain is available, but constructing such a domain for new equipment usually requires substantial manual modeling \cite{kaelbling2011hierarchical, srivastava2014combined}. \emph{MaCoPlanner} likewise requires a shared IR schema and verification machinery; its distinction is that device-specific task knowledge is compiled from equipment manuals rather than manually authored task by task.
In contrast, LVLMs can convert natural-language instructions into structured task sequences and thus improve planning automation \cite{fu2024can}. Recent works have shown the potential of large language models (LLMs) in high-level task decomposition, particularly for long-horizon sequential manipulation \cite{kannan2024smart, zhou2024isr, wang2024llm}. In robotic manipulation, vision--language models (VLMs) further compensate for the lack of visual grounding in LLMs and improve environmental understanding for planning \cite{shirai2024vision}. Some recent frameworks further combine traditional planning structures with LVLMs to improve reliability and execution accuracy \cite{yang2025guiding, ao2025llm, ni2024grid}. Despite these advances, planning for complex industrial equipment remains difficult. Such tasks rely heavily on domain knowledge, including operating procedures, safety regulations, and equipment principles. Existing LVLM-based methods still have limited ability to capture such knowledge, especially functional logic and procedural dependency. Therefore, the generated plans may be technically infeasible or unsafe.

\subsection{Task-Level Safety for Robotic Planning}

Existing robotic safety research has mainly focused on the motion or trajectory level, where collision avoidance and constrained control have been extensively studied \cite{wang2024deep,he2024agile}. In contrast, task-level safety has received much less attention. In traditional closed-world task modeling, task objects, actions, and preconditions are explicitly defined offline, so task execution is naturally restricted within predefined safe boundaries. However, this assumption becomes weak in open-world and natural-language-driven robotic systems based on LVLMs, leading to a clear safety gap at the task-planning level. 

Recent studies have gradually moved safety enforcement upstream into the planning stage. One line of work formalizes natural-language safety rules into linear temporal logic (LTL) and performs runtime monitoring to ensure verifiability and compliance \cite{wu2025selp, yang2024plug}. In contrast to runtime monitoring, the present work uses formal constraints to verify and repair candidate task plans before execution. Another line introduces multi-LLM collaboration or separates task planning from safety planning to reduce conflicts and missing dependencies in long-horizon tasks \cite{khan2025safety}. A third line filters unsafe task goals before decomposition, preventing unsafe targets from entering the execution pipeline \cite{obi2025safeplan}. Although these methods improve safety to different degrees, they still face limitations in cross-device transfer, interpretability, and execution-time anomaly handling. SafeAgentBench further demonstrates safety failures in household embodied tasks \cite{yin2024safeagentbench}; because its task setting differs from industrial panel operation, we use it to motivate the broader embodied-agent safety problem rather than as a directly compatible evaluation benchmark.

Overall, the remaining gap addressed in this work is the use of device manuals to support task-level planning and pre-execution procedural/state verification for industrial panel operation.

\section{MaCoPlanner: Task Planning from Compiled Equipment Manuals with Proactive Safety Verification}

\subsection{Overview of MaCoPlanner}

To address the limited use of manual knowledge and the difficulty of task-level verification in autonomous robotic industrial panel operation, this study proposes \emph{MaCoPlanner}, a framework built around task-planning design choices tailored to industrial panel operation. First, heterogeneous equipment manuals are compiled into machine-usable task knowledge rather than used only as raw prompt context, allowing device-specific procedures, operating conditions, and safety rules to directly support planning. Second, task-relevant evidence is selected according to the current device state to condition VLM-based plan generation, preserving the flexibility of foundation models while incorporating equipment-specific knowledge. Third, verification is placed before physical execution so that generated plans can be checked and repaired before being issued to the robot.

\begin{figure*}[!t]\centering
	\makebox[\linewidth][c]{\includegraphics[width=\textwidth]{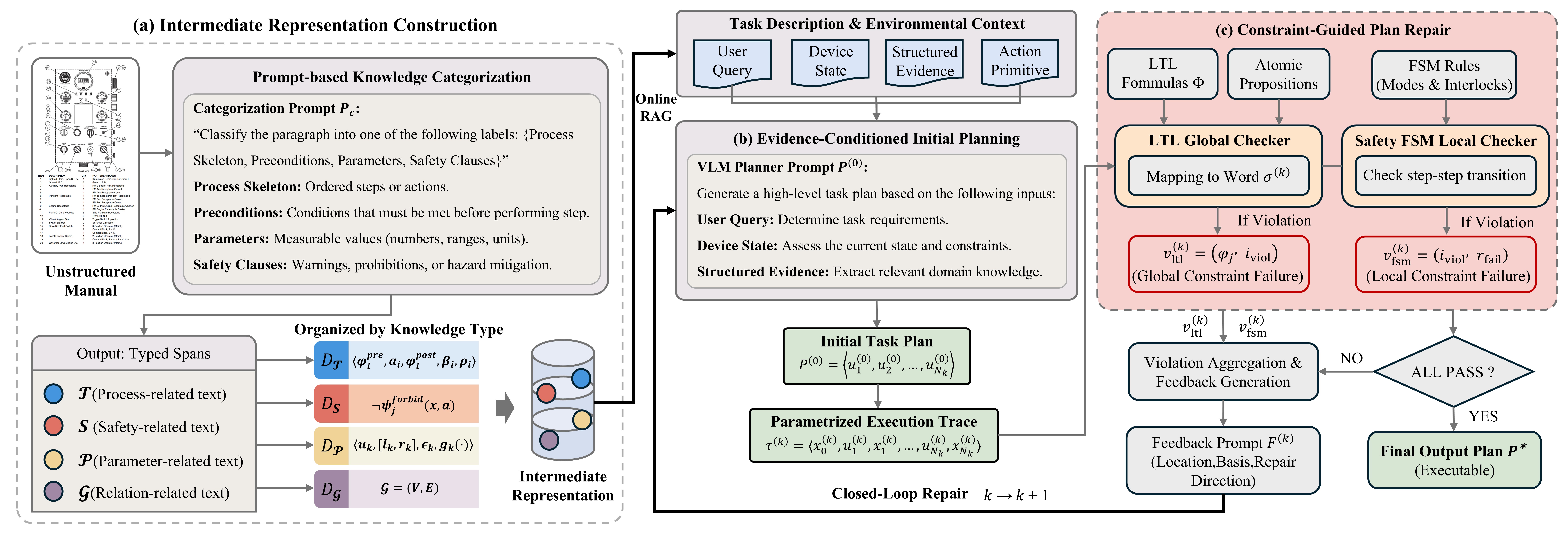}}
	\caption{Planning workflow of \emph{MaCoPlanner}. (a) Manual knowledge compilation converts heterogeneous manuals into a structured IR that preserves procedural, safety, parameter, and relation knowledge. (b) Evidence-conditioned initial planning generates an initial task plan from the user query, device state, and structured evidence. (c) The proactive safety verification mechanism evaluates and iteratively refines the generated plan under procedural and state-dependent constraints before physical execution until a verified symbolic plan is obtained or the refinement budget is exhausted, in which case the task is rejected.}
\label{tu2_overview}
\end{figure*}

As shown in Figure~\ref{tu2_overview}, the compiled knowledge and current device state are used to retrieve task-relevant evidence and generate a candidate sequence of parameterized action primitives. The candidate plan is then symbolically rolled out, with LTL and the Safety FSM checking it from global and local perspectives, respectively. Detected violations are returned to the VLM as targeted feedback for replanning, while a plan that passes both checks is returned as the verified symbolic task sequence. These design choices connect manual-derived knowledge, flexible plan generation, and proactive verification within a unified task-planning process.

\subsection{Compiled Manual Knowledge for Task Planning}\label{subsec:compiled_manual_knowledge}

To address knowledge compilation and retrieval for task planning from equipment manuals, this subsection details how \emph{MaCoPlanner} organizes heterogeneous equipment manuals and retrieves task-relevant evidence. The key idea is to compile and retrieve manual knowledge according to its functional type, allowing different types of knowledge to be represented and accessed separately. This type-aware organization reduces interference among heterogeneous manual content and supports more targeted evidence retrieval. Given a user instruction and the current device state, typed retrieval and evidence gating are then used to select relevant and applicable evidence for subsequent task planning.

\begin{figure}[!t]\centering
	\makebox[\linewidth][c]{\includegraphics[width=5cm]{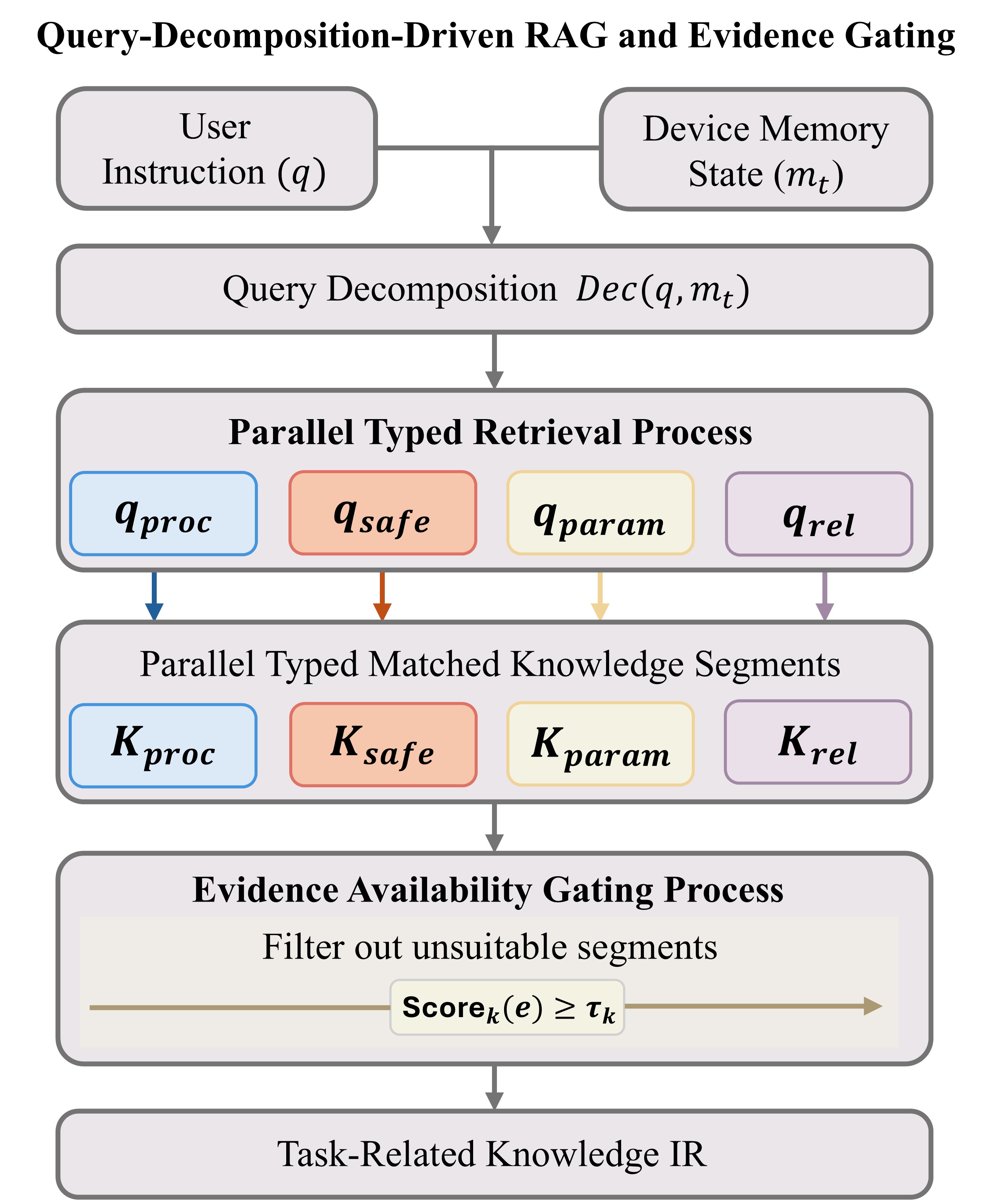}}
	\caption{Query-decomposition-driven retrieval and evidence gating pipeline.}
\label{tu3_overview}
\end{figure}

\subsubsection{Intermediate Representation Construction for Task Planning}\label{subsubsec:ir_construction}
Industrial operation manuals contain dense procedural, safety, parameter, and relation knowledge that is difficult to preserve through direct retrieval over long unstructured text. To address this issue, as illustrated in Figure~\ref{tu2_overview}(a), an offline knowledge compilation method is introduced to map the manual collection $D=\{d_1,\ldots,d_{|D|}\}$ into a typed IR space that is retrievable, traceable, and directly usable for reasoning. Here, the IR is not a latent embedding space, but a typed and compiled task-oriented knowledge representation, where normalized procedural steps, safety constraints, parameter specifications, and control-state relations are explicitly encoded for downstream retrieval, planning, and verification. Formally, the compilation operator $\operatorname{Compile}(\cdot)$ transforms the manual collection into
\begin{equation}\label{eq:ir_compile}
\mathcal{I}=\operatorname{Compile}(D)=(\mathcal{T},\mathcal{S},\mathcal{P},\mathcal{G}),
\end{equation}
where $\mathcal{I}$ denotes the compiled IR, and $\mathcal{T}$, $\mathcal{S}$, $\mathcal{P}$, and $\mathcal{G}$ denote the process skeleton set, safety/compliance clause set, parameter specification set, and structural relation graph, respectively. A provenance mapping $\ell$ associates each compiled item $i$ with its source manual and text span, $\ell(i)=(d,\omega)$, where $d\in D$ and $\omega$ identifies the corresponding source span. In implementation, $\operatorname{Compile}(\cdot)$ includes four stages: manual chunking and clause extraction, LLM-assisted knowledge-type classification, template-based normalization, and cross-item linking, with source back-links preserved through $\ell$. Hence, \eqref{eq:ir_compile} denotes a manual-to-structured-knowledge transformation rather than simple retrieval chunking. 

Missing or ambiguous fields are explicitly recorded during normalization. Since classification and rule translation are model-assisted, compilation errors may still occur; therefore, the validity of formal verification ultimately depends on the fidelity of the manual-to-IR compilation.

Within $\mathcal{T}$, each procedure is decomposed into step nodes $\{s_i\}$ with explicit execution semantics:
\begin{equation}
s_i=\langle \phi_i^{pre},a_i,\phi_i^{post},\beta_i,\rho_i\rangle.
\end{equation}
Here, $A$ denotes the action type set, $\phi_i^{pre}$ is a decidable predicate over the current system state $x$, $a_i\in A$ denotes an action primitive together with its parameterization, and $\phi_i^{post}$ denotes the acceptance criterion verifiable from observation $o$. The term $\beta_i$ specifies the guard condition for exception-triggered transition, and $\rho_i$ constrains the retry policy. This representation converts procedural text into executable step structures, so that prerequisites, expected outcomes, and exception branches are all explicitly available during planning.

Each clause in $\mathcal{S}$ is further compiled into a trigger-forbidden constraint:
\begin{equation}
k_j:\phi_j^{trig}(x)\Rightarrow \neg \psi_j^{forbid}(x,a),
\end{equation}
where $\phi_j^{trig}$ denotes the triggering condition and $\psi_j^{forbid}$ denotes the prohibited action or action set. In this way, safety knowledge becomes directly callable planning constraints rather than implicit textual reminders.

To handle parameter-intensive specifications with unit variations, $\mathcal{P}$ normalizes each parameter item into
\begin{equation}
p_k=\langle u_k,[l_k,r_k],\epsilon_k,g_k(\cdot)\rangle,
\end{equation}
where $u_k$ is the unit, $[l_k,r_k]$ is the allowable interval, $\epsilon_k$ is the tolerance, and $g_k(o)$ is the observation-based extraction and verification function. This representation enables parameter-consistent and executable action generation. Finally, the spatial layout and logical coupling among controls are compiled into the structural graph $\mathcal{G}$:
\begin{equation}
\mathcal{G}=(V,E),
\end{equation}
where $V$ contains controls, steps, parameters, and state predicates, and $E$ encodes spatial adjacency, affiliation, and dependency relations extracted from layout descriptions, control-step references, and device logic statements in the manuals. This graph provides structural priors for subsequent typed retrieval and consistency checking.

\subsubsection{Query-Decomposition-Driven Retrieval and Evidence Gating}
\label{subsubsec:query_decomp_retrieval}

In the online stage, as shown in Figure~\ref{tu3_overview}, retrieval is conditioned on the user instruction $q$ and the device memory state $m_t$. Since high-level instructions are often abstract while relevant knowledge is distributed across procedures, safety clauses, parameters, and control relations, direct single-query retrieval from unstructured text may miss critical constraints or introduce irrelevant evidence. To address this issue, a retrieval decomposition operator is introduced to map $(q,m_t)$ into typed subqueries constrained by IR slots.

Let the typed index collection be
\begin{equation}\label{eq:index_set}
\mathcal{K}=\{K_{proc},K_{safe},K_{param},K_{rel}\},
\end{equation}
where $\mathcal{K}$ denotes the set of typed knowledge indices, and $K_{proc}$, $K_{safe}$, $K_{param}$, and $K_{rel}$ store process structures, safety clause specifications, parameter objects, and graph-based relation evidence, respectively. Based on the compiled IR in \eqref{eq:ir_compile}, the online retriever organizes manual knowledge through the typed indices in \eqref{eq:index_set}. The retrieval decomposition is then defined as
\begin{equation}\label{eq:retrieval_dec}
Z=Dec(q,m_t;\mathcal{I})=\bigcup_{c\in\{\mathrm{proc,safe,param,rel}\}} Z_c,
\end{equation}
where $Z_c=\{z_{(c,1)},\ldots,z_{(c,n_c)}\}$ denotes the typed retrieval demands aligned with category $c$.

In implementation, $Dec(\cdot)$ is a prompt-based structured decomposition step constrained by the four IR slots and followed by schema post-processing.

It reads the user instruction and current device state and emits zero or more typed demands for procedure, safety, parameter, and relation evidence. Procedure demands identify the required task progression and prerequisites, safety demands identify relevant constraints, parameter demands identify required setpoints or ranges, and relation demands identify controls and their associated relations.

Malformed or out-of-schema fields are rejected or normalized during post-processing. Implementation details and fixed settings for query decomposition, retrieval, and evidence gating are summarized in \ref{app:reproducibility}.

For each atomic demand in \eqref{eq:retrieval_dec}, a typed query vector is constructed as
\begin{equation}\label{eq:typed_query}
q_{(c,i)}=f_{enc}(q\oplus z_{(c,i)}\oplus m_t;\theta_{\mathrm{enc}}), 
\quad c\in\{\mathrm{proc,safe,param,rel}\},
\end{equation}
where $\oplus$ denotes structured concatenation, $f_{enc}(\cdot)$ denotes the dense query encoder, and $\theta_{\mathrm{enc}}$ denotes its shared encoder parameters. Type specificity is introduced through the query templates or prefixes associated with each knowledge type, allowing the same instruction to be represented differently while using the same dense retriever. Retrieval is then performed in parallel as
\begin{equation}\label{eq:parallel_retrieval}
E_{(c,i)}=TopL(K_c,q_{(c,i)}), 
\qquad 
E_c=\bigcup_{i=1}^{n_c} E_{(c,i)}.
\end{equation}
Here, $TopL(\cdot)$ denotes top-$L$ nearest-neighbor retrieval in the corresponding typed vector index.

The same embedding, similarity, and retrieval-depth settings are fixed across all controlled retrieval conditions.

The chain \eqref{eq:retrieval_dec}--\eqref{eq:typed_query}--\eqref{eq:parallel_retrieval} makes the retrieval dependency explicit: the instruction and memory are first decomposed into typed demands, which are then encoded and routed to the corresponding indices in parallel. This type-aware retrieval strategy improves access to task-relevant evidence while reducing interference from unrelated manual content.

To characterize constraint completeness, let $R(q,m_t)$ denote the set of planning requirements and let $cov(e)\subseteq R$ denote the subset covered by evidence item $e$. The evidence coverage ratio is defined as
\begin{equation}\label{eq:coverage}
Cover(E,R)=\frac{1}{|R|}\sum_{r\in R}\mathbf{1}[\exists e\in E:r\in cov(e)].
\end{equation}

Equation~\eqref{eq:coverage} is used as an empirical requirement-coverage metric rather than an analytical guarantee and is evaluated under the fixed corpus and context configuration used in Section~\ref{subsec:representation_scalability}.

After retrieval, evidence-availability gating is applied to remove semantically related but inapplicable evidence. The basic gating score is defined as
\begin{equation}
Score(e,q_{(c,i)},m_t)
=
\alpha_c\,Sim(e,q_{(c,i)})
+
(1-\alpha_c)\,App(e,m_t),
\end{equation}
where $Sim(\cdot)$ denotes dense semantic similarity and $App(\cdot)$ denotes a rule-based applicability term that checks whether the evidence is consistent with the current device mode, target control, and state or parameter predicates recorded in $m_t$ and $\mathcal{I}$.

Let $\mathcal{H}_e\equiv\mathcal{H}(e)$ be the set of mode, control, state, and parameter predicates explicitly attached to evidence item $e$, let $g_h(e,m_t)\in\{0,1\}$ indicate whether check $h$ is satisfied, and let $w_h>0$ be its predefined weight. Let $\mathcal{H}_{\mathrm{hard}}(e)\subseteq\mathcal{H}_e$ contain the predicates whose contradiction makes the evidence inapplicable, and let $C_h\equiv C_h(e,m_t)$ be the corresponding hard-conflict indicator: $C_h=1$ if and only if $g_h(e,m_t)=0$ for some $h\in\mathcal{H}_{\mathrm{hard}}(e)$. For nonempty $\mathcal{H}_e$, the weighted predicate-satisfaction score $\bar{g}_w(e,m_t)$ and the applicability term are computed as
\begin{equation}\label{eq:applicability_score}
\begin{aligned}
\bar{g}_w(e,m_t)
&=
\frac{\sum_{h\in\mathcal{H}_e}w_h g_h(e,m_t)}
{\sum_{h\in\mathcal{H}_e}w_h},\\[3pt]
App(e,m_t)
&=
\begin{cases}
0, & C_h=1,\\[2pt]
\bar{g}_w(e,m_t), & C_h=0,\ \mathcal{H}_e\neq\varnothing,\\[2pt]
1, & \mathcal{H}_e=\varnothing.
\end{cases}
\end{aligned}
\end{equation}
Checks not specified for an evidence item are omitted rather than treated as failures. The score and weights are predefined rather than learned; the deployed check inventory and weights are reported in \ref{app:reproducibility}.

For any $e\in E_c$, its maximal availability score is then defined as
\begin{equation}
Score_c(e;q,m_t)
=
\max_{i\in\{1,\ldots,n_c\}}
Score(e,q_{(c,i)},m_t),
\end{equation}
and the gated evidence set is obtained as
\begin{equation}
\hat{E}_c
=
\{e\in E_c \mid C_h(e,m_t)=0 \ \land\ Score_c(e;q,m_t)\ge \tau_c\}.
\end{equation}

The current implementation applies the hard-conflict test as a veto before thresholding and uses $\alpha_c=\alpha=0.70$ and one shared cutoff $\tau_c=\tau=0.55$ for all evidence types. The value $\alpha=0.70$ gives semantic relevance the larger contribution while reserving 30\% of the combined score for state applicability. The cutoff $\tau=0.55$ is a conservative near-midrange gate: weakly supported evidence is removed, whereas hard contradictions are rejected independently by the veto and cannot pass by score compensation. All applicability predicates use equal weights. These settings were fixed before benchmark evaluation and held constant across all retrieval variants; they are reported for reproducibility rather than claimed as generally optimal.

The final evidence set is then written as
\begin{equation}\label{eq:gated_evidence_final}
E_t=\hat{E}=\bigcup_c \hat{E}_c,
\end{equation}
and \eqref{eq:gated_evidence_final} is injected into the VLM context as explicit planning evidence and constraints.

\subsection{Proactive Safety Verification Mechanism for Task Planning}\label{subsec:proactive_verification}
To verify task-level procedural order and state-transition legality before execution, this subsection presents \emph{MaCoPlanner}'s proactive safety verification mechanism for iterative plan improvement. The key idea is to treat safety not as a post hoc filtering step, but as a pre-execution constraint-guided process in which generated plans are evaluated, diagnosed, and refined before physical action. Specifically, the initial plan is generated under retrieved evidence and then assessed under two complementary types of constraints, namely procedural constraints across multi-step task progression and state-dependent constraints induced by device modes and interlocks. Detected violations are converted into structured feedback for targeted replanning, so that safety requirements become explicit refinement signals during iterative planning.

\subsubsection{Evidence-Conditioned Initial Planning}\label{subsubsec:evidence_conditioned_planning}
Instead of directly injecting the full manual into the VLM prompt, as shown in Figure~\ref{tu2_overview}(b), the initial plan is generated under retrieved task-relevant evidence. Specifically, the structured retrieval module introduced in Section~\ref{subsec:compiled_manual_knowledge} extracts evidence from the manual and device state, yielding an evidence set $E_t$ aligned with the user instruction $q$ and current device memory state $m_t$. The VLM planner first generates the initial plan as
\begin{equation}\label{eq:init_plan}
P^{(0)}=\pi_{\theta_{\mathrm{pln}}}(q,m_t,E_t).
\end{equation}
Here, $\pi_{\theta_{\mathrm{pln}}}(\cdot)$ denotes a prompted VLM planner with parameters $\theta_{\mathrm{pln}}$. It takes the user goal, the summarized device state, and the gated manual evidence as input, and outputs a high-level primitive sequence over a fixed action library (e.g., move-to-control, press, rotate, toggle, and verify).

For subsequent proactive safety verification, the plan in \eqref{eq:init_plan} is rewritten as a sequence of parameterized action primitives:
\begin{equation}\label{eq:plan_sequence}
P^{(k)}=\langle u_1^{(k)},u_2^{(k)},\ldots,u_{N_k}^{(k)} \rangle,
\end{equation}
\begin{equation}\label{eq:plan_action}
u_i^{(k)}=(a_i^{(k)},p_i^{(k)}), \qquad a_i^{(k)}\in A.
\end{equation}
Here, $p_i^{(k)}$ denotes the action parameters.

Given \eqref{eq:plan_sequence}--\eqref{eq:plan_action}, the device state is represented by a symbolic state $x\in X$, derived from observation, memory, and IR rules. Here, $X$ denotes the symbolic state space. Under the abstract transition operator $\hat{f}$, the symbolic state evolution is written as
\begin{equation}\label{eq:state_evolution}
x_0^{(k)}=Abs(m_t), \qquad
x_{i+1}^{(k)}=\hat{f}(x_i^{(k)},u_{i+1}^{(k)}), \ i=0,\ldots,N_k-1,
\end{equation}
where $Abs(\cdot)$ abstracts the device memory into a symbolic state consisting of slots such as power state, operation mode, direction state, alarm flags, and key setpoints, and $\hat{f}(\cdot)$ denotes the abstract transition operator generated from compiled preconditions/postconditions and the device safety finite-state machine (FSM). 
Equation~\eqref{eq:state_evolution} represents a deterministic symbolic rollout rather than the physical device dynamics. Therefore, verification guarantees plan legality only with respect to the encoded symbolic state and transition model; errors in state estimation or model abstraction may cause the actual device state to differ from the predicted rollout.
 The state rollout in \eqref{eq:state_evolution} then induces the finite execution trace
\begin{equation}\label{eq:execution_trace}
\xi^{(k)}=\langle x_0^{(k)},u_1^{(k)},x_1^{(k)},\ldots,u_{N_k}^{(k)},x_{N_k}^{(k)} \rangle.
\end{equation}
Equation~\eqref{eq:execution_trace} provides the explicit object that is checked and repaired in the subsequent verification mechanism.

\subsubsection{Constraint-Guided Verification and Plan Repair}\label{subsubsec:constraint_guided_repair}
Industrial safety rules, as shown in Figure~\ref{tu2_overview}(c), usually involve two complementary forms of constraints: procedural constraints that govern temporal correctness across multiple steps, and state-dependent constraints that govern local legality under device modes and interlocks. Accordingly, the proposed method evaluates the generated plan from these two perspectives and converts detected violations into localized feedback for refinement. In implementation, the first class is instantiated as finite-trace LTL (LTLf) formulas compiled from the procedure and safety IR, while the second class is instantiated as a Safety FSM whose transitions encode legal mode/interlock changes.

Based on the execution trace in \eqref{eq:execution_trace}, procedural constraint checking evaluates global temporal safety. Let $AP$ denote the set of atomic propositions, and let the labeling function $\widetilde{L}:X\times A\rightarrow 2^{AP}$ label each action together with its pre-action symbolic state. The proposition sequence induced by \eqref{eq:execution_trace} is
\begin{equation}
\sigma^{(k)}=\langle \widetilde{L}(x_0^{(k)},u_1^{(k)}),\ldots,\widetilde{L}(x_{N_k-1}^{(k)},u_{N_k}^{(k)}),L_T(x_{N_k}^{(k)})\rangle,
\end{equation}
where $L_T$ labels the terminal symbolic state and contains no action event. For example, the panel-OFF rule in Table~\ref{tab:appendix_ltl_examples} requires the load to be off in the symbolic state immediately before the OFF-button action, rather than allowing it to be switched off afterward.

Each finite candidate plan is checked using LTLf. For a failed local precondition or FSM transition, $i_{\mathrm{viol}}$ is the first failing action. For a prefix-closed LTLf safety formula, it is the earliest prefix with no satisfying continuation. If a non-safety eventuality is still unsatisfied at the terminal state, the plan is invalid and $i_{\mathrm{viol}}=N_k$.

Given the safety formula set $\Phi$, the procedural check is written as
\begin{equation}\label{eq:proc_check}
\sigma^{(k)} \models \Phi \Longleftrightarrow \forall \phi \in \Phi,\ \sigma^{(k)} \models \phi.
\end{equation}
If \eqref{eq:proc_check} fails, the system returns
\begin{equation}\label{eq:proc_violation}
v_{proc}^{(k)}=(\phi_j,i_{viol}),
\end{equation}
where $\phi_j$ is the violated formula and $i_{viol}$ is the earliest violation index.

State-dependent constraint checking focuses on local transition legality. Operationally, each primitive in $P^{(k)}$ is replayed on the Safety FSM generated from device operating logic; if a transition is illegal (for example, a direction switch that skips the OFF state), the checker immediately records the first failure. If all steps are legal, it returns $\top$; otherwise, it returns
\begin{equation}\label{eq:state_violation}
v_{state}^{(k)}=(i_{viol},r_{fail}),
\end{equation}
where $r_{fail}$ is the readable failure reason. When either \eqref{eq:proc_violation} or \eqref{eq:state_violation} is triggered, the outputs are merged into a structured violation report $V^{(k)}$ and converted into a feedback prompt $F^{(k)}$. The refined plan is then generated as
\begin{equation}
P^{(k+1)}=\pi_{\theta_{\mathrm{pln}}}(q,m_t,E_t,F^{(k)}),
\end{equation}
expanded into a new trace $\xi^{(k+1)}$, and re-evaluated until all constraints are satisfied or the maximum number of refinement attempts is reached.

Only plans passing both checkers are accepted for execution. We set $K_{\max}=3$ to provide sufficient opportunity for plan repair while limiting repeated model calls and online latency. If no admissible plan is obtained within this budget, the task is rejected without issuing physical actions. Algorithm~\ref{alg:macoplanner_online} summarizes the complete procedure from evidence retrieval and plan generation to verification, refinement, and the final decision.

Verification remains conditional on specification fidelity and symbolic-state correctness.

\begin{algorithm}[t]
\caption{Online planning from compiled equipment-manual knowledge with fail-closed pre-execution verification in \emph{MaCoPlanner}.}
\label{alg:macoplanner_online}
\begin{algorithmic}[1]
\Require User instruction $q$, current device state $m_t$, compiled IR $\mathcal{I}$, typed indices $\mathcal{K}$
\Ensure Verified symbolic plan $P^{*}$ or an explicit rejection
\State $Z \gets Dec(q,m_t;\mathcal{I})$
\For{each type $c\in\{\mathrm{proc,safe,param,rel}\}$}
    \State construct typed subqueries $\{q_{(c,i)}\}$ from $Z_c$
    \State retrieve $E_{(c,i)} \gets TopL(K_c,q_{(c,i)})$ for $i=1,\ldots,n_c$
    \State gate retrieved evidence and obtain $\hat{E}_c$
\EndFor
\State $E_t \gets \bigcup_c \hat{E}_c$
\State $P^{(0)} \gets \pi_{\theta_{\mathrm{pln}}}(q,m_t,E_t)$
\State $verified \gets \mathrm{false}$
\For{$k=0$ to $K_{\max}$}
    \State initialize symbolic state $x_0^{(k)} \gets Abs(m_t)$ and rollout $\xi^{(k)}$ using $\hat{f}$
    \State evaluate finite-trace procedural constraints $\Phi$ on $\xi^{(k)}$
    \State evaluate local legality by replaying $P^{(k)}$ on the Safety FSM
    \If{no violation is detected}
        \State $P^{*} \gets P^{(k)}$; $verified\gets\mathrm{true}$; \textbf{break}
    \ElsIf{$k < K_{\max}$}
        \State build violation report $V^{(k)}$ and feedback prompt $F^{(k)}$
        \State $P^{(k+1)} \gets \pi_{\theta_{\mathrm{pln}}}(q,m_t,E_t,F^{(k)})$
    \EndIf
\EndFor
\If{$verified=\mathrm{false}$}
    \State \Return \textsc{RejectTask}; do not invoke grounding or physical execution
\EndIf
\State \Return $P^{*}$
\end{algorithmic}
\end{algorithm}

\section{Manual-Guided Grounding and Device Memory}\label{sec:execution_interface}\label{subsec:grounding_memory}

\begin{figure*}[!t]\centering
	\makebox[\linewidth][c]{\includegraphics[width=13cm]{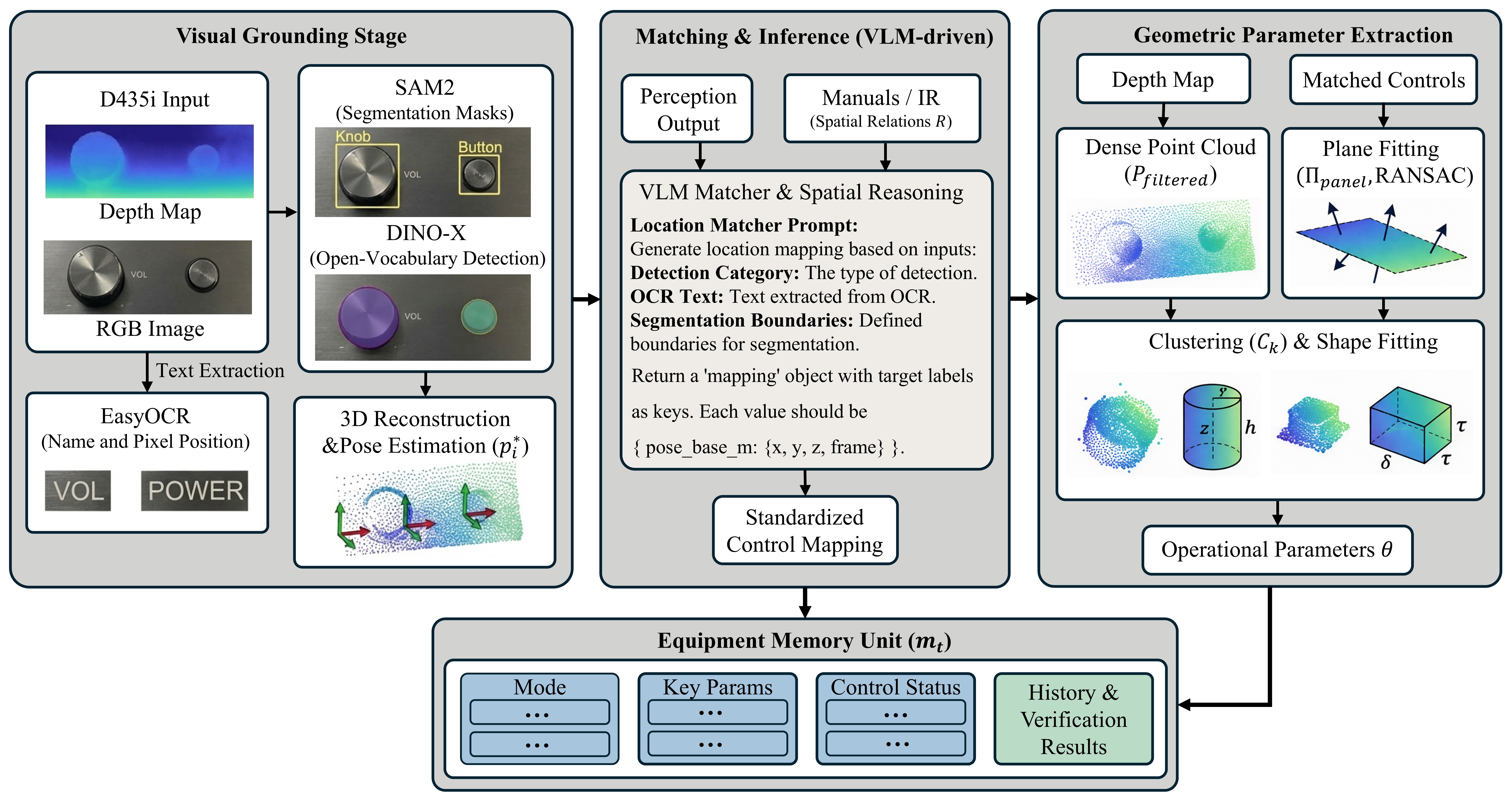}}
	\caption{Illustration of manual-guided grounding and device memory. Multimodal perception and geometric parameter extraction are used to align symbolic controls with physical instances and support state-consistent execution.}
\label{tu4_overview}
\end{figure*}

After \emph{MaCoPlanner} returns a verified symbolic task plan, physical operation requires a separate execution interface that maps symbolic controls to physical panel instances and updates the device state from observed outcomes. As shown in Figure~\ref{tu4_overview}, this section describes the manual-guided grounding and device-memory interface used for this purpose.
The grounding process is formulated as a multimodal matching problem jointly constrained by semantic consistency and spatial-relation consistency, and further outputs operation-ready parameters for execution and verification. The device memory records modes, key parameters, control states, and verification outcomes after physical operation, providing the updated state input required by subsequent \emph{MaCoPlanner} planning calls.

At the visual grounding stage, SAM2~\cite{ravi2024sam} is used to generate pixel-level segmentation masks, DINO-X~\cite{ren2024dino} performs open-vocabulary detection, and optical character recognition (OCR) extracts panel labels as semantic anchors. Through depth reconstruction, pixel-space boundaries and keypoints are mapped to the robot base frame, yielding the observed pose $p_i^{*}$ of each candidate control instance. To improve matching robustness, control aliases and layout priors from manuals/IR are incorporated as a spatial-relation vocabulary and its instantiated relation set:
\begin{equation}\label{eq:ground_rel_vocab}
\mathcal{R}_{s}=\{\text{left\_of},\text{right\_of},\text{above},\text{below},\text{near},\ldots\},
\end{equation}
\begin{equation}\label{eq:ground_rel_inst}
\mathcal{R}=\{(a,r,b)\mid r\in \mathcal{R}_{s}\}.
\end{equation}
Here, $\mathcal{R}_{s}$ denotes the predefined spatial-relation vocabulary, and $\mathcal{R}$ denotes the relation triplets instantiated for the current panel. Detection-class consistency and OCR/control-alias overlap first produce a coarse candidate-assignment set $\mathcal{C}$. For regularly arranged control groups, each assignment $c\in\mathcal{C}$ is ranked using the deterministic geometric and relation-consistency score
\begin{equation}\label{eq:ground_match}
\begin{aligned}
S_{\mathrm{match}}(c)
&=\sum_{i=1}^{n}\|p_i^{(c)}-p_i^{*}\|^2
+\lambda\sum_{(a,r,b)\in \mathcal{R}}
g_r\!\left(p_a^{(c)},p_b^{(c)}\right),\\
c^{*}
&=\arg\min_{c\in\mathcal{C}}S_{\mathrm{match}}(c).
\end{aligned}
\end{equation}
Here, $p_i^{(c)}$ is the control pose assigned by candidate $c$, $p_i^{*}$ is the observed pose, $g_r(\cdot)$ penalizes violations of spatial relation $r$, and lower scores indicate better consistency. Equation~\eqref{eq:ground_match} is a candidate-ranking rule, not a learned training objective. The deployed relation weight is $\lambda=0.5$, which keeps relation consistency as a secondary correction to the pose term rather than allowing it to dominate class/OCR and geometric evidence. After deterministic filtering, candidates with $S_{\mathrm{match}}\leq0.65$ are retained; this permissive boundary avoids over-pruning plausible controls, while the VLM is invoked only for remaining ties or semantic ambiguity. Both values were fixed before evaluation and used unchanged across the five camera viewpoints; they are implementation settings rather than transferable optima. Geometric thresholds and clustering parameters are listed in \ref{app:reproducibility}.

Spatial localization alone is insufficient for panel operation, since the robot also requires operation-critical parameters such as knob radius, button stroke, and triggering threshold. Unlike general grasp methods such as AnyGrasp~\cite{fang2023anygrasp}, our goal is to estimate physical parameters directly usable for functional execution and step-level verification. To this end, the grounding result of each control is extended from pose $p$ to $\langle p,\boldsymbol{\eta}_{\mathrm{op}}\rangle$, where $\boldsymbol{\eta}_{\mathrm{op}}$ denotes the operation parameter vector. Stereo depth images are first processed by FoundationStereo~\cite{wen2025foundationstereo} to recover the raw point cloud and its denoised version:
\begin{equation}\label{eq:pointcloud_raw}
P_{\mathrm{raw}}=\left\{p_i\in \mathbb{R}^{3}\mid i=1,\ldots,N\right\},
\end{equation}
\begin{equation}\label{eq:pointcloud_filtered}
P_{\mathrm{filtered}}=\operatorname{Denoise}\!\left(P_{\mathrm{raw}}\right),
\end{equation}
where $N$ is the number of recovered 3D points and $\operatorname{Denoise}(\cdot)$ denotes outlier removal and local smoothing before geometric fitting. A panel reference plane $\Pi_{\mathrm{panel}}=\{n_{\mathrm{panel}},d_{\mathrm{panel}}\}$ is then estimated from $P_{\mathrm{filtered}}$ by
\begin{equation}\label{eq:panel_plane_fit}
(n_{\mathrm{panel}},d_{\mathrm{panel}})=\arg\min_{\|n\|_2=1,\,d}\sum_{p_i\in P_{\mathrm{filtered}}}\mathcal{L}_{\mathrm{rob}}\!\left(|n^{T}p_i-d|\right),
\end{equation}
where $n_{\mathrm{panel}}$ and $d_{\mathrm{panel}}$ denote the plane normal and offset, respectively, and $\mathcal{L}_{\mathrm{rob}}(\cdot)$ is a robust loss. Using the fitted plane in \eqref{eq:panel_plane_fit}, the point cloud is aligned to the panel coordinate system by
\begin{equation}\label{eq:panel_alignment}
p_{\mathrm{panel}}=\mathbf{R}_{\mathrm{align}}\bigl(p_{\mathrm{world}}-o_{\mathrm{panel}}\bigr),
\end{equation}
where $\mathbf{R}_{\mathrm{align}}$ denotes the alignment rotation, $p_{\mathrm{world}}$ denotes the original 3D point in the world frame, and $o_{\mathrm{panel}}$ denotes the chosen panel-frame origin. Control candidate clusters are then extracted by DBSCAN-style density clustering:
\begin{equation}\label{eq:panel_cluster}
\{C_k\}_{k=1}^{N_{\mathrm{clust}}}=\operatorname{DBSCAN}\!\left(P_{\mathrm{panel}};\epsilon_{\mathrm{db}},\mathrm{MinPts}\right),
\end{equation}
where $C_k$ denotes the $k$th density-connected candidate cluster, $\epsilon_{\mathrm{db}}$ is the DBSCAN neighborhood radius, $\mathrm{MinPts}$ is the minimum neighborhood size, and $N_{\mathrm{clust}}$ is the number of clusters returned by the density-clustering procedure. Shape-based filtering is then applied to the resulting clusters to obtain the optimal cluster $C_{\mathrm{optimal}}$. For a knob cluster $C_{\mathrm{knob}}$, a cylindrical/disk model is fitted to estimate the center $c$, radius $r$, and height $h$:
\begin{equation}\label{eq:knob_fit}
\min_{r,h,c}\sum_{p\in C_{\mathrm{knob}}}\left(\|p_{xy}-c_{xy}\|-r\right)^2+g_{\mathrm{thick}}(p_z,h),
\end{equation}
where $p_{xy}$ and $c_{xy}$ denote the planar coordinates of a point and the fitted center, respectively, $p_z$ denotes the point height in the aligned panel frame, and $g_{\mathrm{thick}}(\cdot)$ constrains height/thickness consistency. For a button cluster $C_{\mathrm{button}}$, local plane/box fitting with movable-margin estimation along the normal direction is used to estimate the pressing stroke $\delta_{\mathrm{button}}$ and triggering threshold $\eta_{\mathrm{button}}$. The resulting operation parameter set is
\begin{equation}\label{eq:operation_params}
\boldsymbol{\eta}_{\mathrm{op}}=\{r_{\mathrm{knob}},\delta_{\mathrm{button}},\eta_{\mathrm{button}},\vartheta_{\mathrm{handle}},\ldots\}.
\end{equation}
The chain \eqref{eq:pointcloud_raw}--\eqref{eq:panel_alignment} converts stereo depth observations into panel-aligned geometric evidence, while \eqref{eq:panel_cluster}--\eqref{eq:operation_params} extracts control-specific execution parameters for downstream manipulation and verification.

Based on the grounding result, the device memory maintains device modes, key parameters, control states, and historical verification results in structured slots, and updates them after each action using observations and step-level completion criteria, yielding the conditioned state $m_t$. This state serves as the condition input for retrieval and evidence gating in Section~\ref{subsec:compiled_manual_knowledge} and for procedural/state-dependent checking in Section~\ref{subsec:proactive_verification}. The memory update reduces state drift but does not eliminate sensing, grounding, stale-state, or unmodeled-dynamics errors; these remain outside the formal guarantee of the task-level verifier.

\begin{table*}[t]

\centering
\caption{Summary of benchmarks, robotic setup, and evaluation metrics.}
\label{tab:exp_settings}
\tiny
\renewcommand{\arraystretch}{0.72}
\setlength{\tabcolsep}{3.5pt}
{
\adjustbox{max width=\textwidth}{
\begin{tabular}{
>{\arraybackslash}p{2.5cm}
>{\arraybackslash}p{3.6cm}
>{\arraybackslash}p{4.0cm}
>{\centering\arraybackslash}p{1.3cm}
>{\centering\arraybackslash}p{1.2cm}
>{\arraybackslash}p{5.2cm}}
\toprule
\tableheaderrow
\textbf{Setting} &
\textbf{Device} &
\textbf{Source} &
\textbf{Scale} &
\textbf{Evaluation Mode} &
\textbf{Reported Metrics} \\
\cmidrule(lr){1-6}

Manual compilation fidelity
& Motor-drive controller simulator / Generator control system / Laser controller / Engine controller
& Operation manuals, safety specifications, parameter specifications
& 350 annotated audit units
& Offline
& \metric{Completeness}, \metric{Semantic Consistency} \\

\cmidrule(lr){1-6}

Retrieval evaluation
& Motor-drive controller simulator / Generator control system / Laser controller / Engine controller
& Annotated evidence aligned with compiled IR
& 400
& Offline
& \metric{Hit@K}, \metric{Recall@K}, \metric{Precision@K}, \metric{F1@K}, \metric{nDCG@K} \\

\cmidrule(lr){1-6}

Task planning and safety evaluation
& Same four domains as retrieval evaluation
& Same source as the retrieval benchmark
& 480
& Symbolic
& \metric{Success Rate}, \metric{Violation Rate}, \metric{Average Refinement Steps} \\

\cmidrule(lr){1-6}

Perception and operation
& Motor-drive controller simulator
& Image observation and depth geometry from five prescribed camera viewpoints
& 5 camera views
& Physical
& \metric{Measurement Error}, \metric{Operation Success Rate} \\

\midrule

Long-horizon execution
& Same as above
& Device memory, manual evidence, state feedback
& 25 repeated trials
& Physical
& \metric{Inference Time}, \metric{Execution Time}, \metric{Completion Rate}, \metric{Safety Trigger Count} \\

\bottomrule
\end{tabular}
}}
\end{table*}

\section{Experiment}
\subsection{Experimental Setup}
\emph{MaCoPlanner} is evaluated through manual-compilation fidelity, retrieval/representation quality, task-planning performance, independently assessed task-level safety, formal-rule/repair analysis, and physical execution. 
A summary of the benchmarks, robotic setup, and evaluation metrics is provided in Table~\ref{tab:exp_settings}. 
The physical platform uses a controller-panel simulator without an attached industrial load, which isolates panel interaction and state-dependent execution from load-driven process dynamics while preserving the planning--verification--grounding--feedback chain examined in this work. 

\subsubsection{Benchmark Construction and Experimental Protocol}
To evaluate \emph{MaCoPlanner} under controlled conditions, we construct two complementary benchmarks for evidence retrieval and high-level task planning, drawing on the benchmark design principles of prior retrieval and safety-aware planning studies~\cite{friel2024ragbench,valmeekam2023planbench,sadhu2025vestabench}.

The first benchmark contains 400 retrieval tasks, each defined by a natural-language instruction and current device state. Reference evidence is independently annotated from the original manuals: annotators identify the complete task-required evidence and the subset applicable to the given device state, with agreement measured before adjudication. Source-level annotations are mapped to compiled items only when evaluating retrieval over the compiled representation. ~\ref{app:benchmark_protocol} details the annotation, blinding, agreement, and adjudication protocol. All retrieval methods use the same candidate pool and $K\in\{8,12,16\}$.

The second benchmark contains 480 high-level panel-operation tasks, including 80 Level-1, 200 Level-2, and 200 Level-3 tasks. Each task provides a structured initial state, a natural-language goal, and an expert reference sequence over the shared action primitives. Task definitions, reference sequences, safety requirements, and control relations are audited against the original manuals. A separate external safety oracle is independently constructed from the same sources and withheld from the planner, verifier, and repair loop. The 400 retrieval tasks and 480 planning tasks are independently constructed for their respective evaluations; they share the same device/manual domains but are not paired one-to-one.

All planner API calls are stateless and task-local. Each task begins with a fresh context, and any subsequent repair request receives only the current task context, candidate plan, and verifier feedback; no conversational memory is shared across tasks. \ref{app:reproducibility} reports the planner model snapshots and request-isolation protocol. All tasks receive equal weight when computing the reported aggregate metrics.

\subsubsection{Robotic Experimental Platform}
\begin{figure}[!t]\centering
	\makebox[\linewidth][c]{\includegraphics[width=6.0cm]{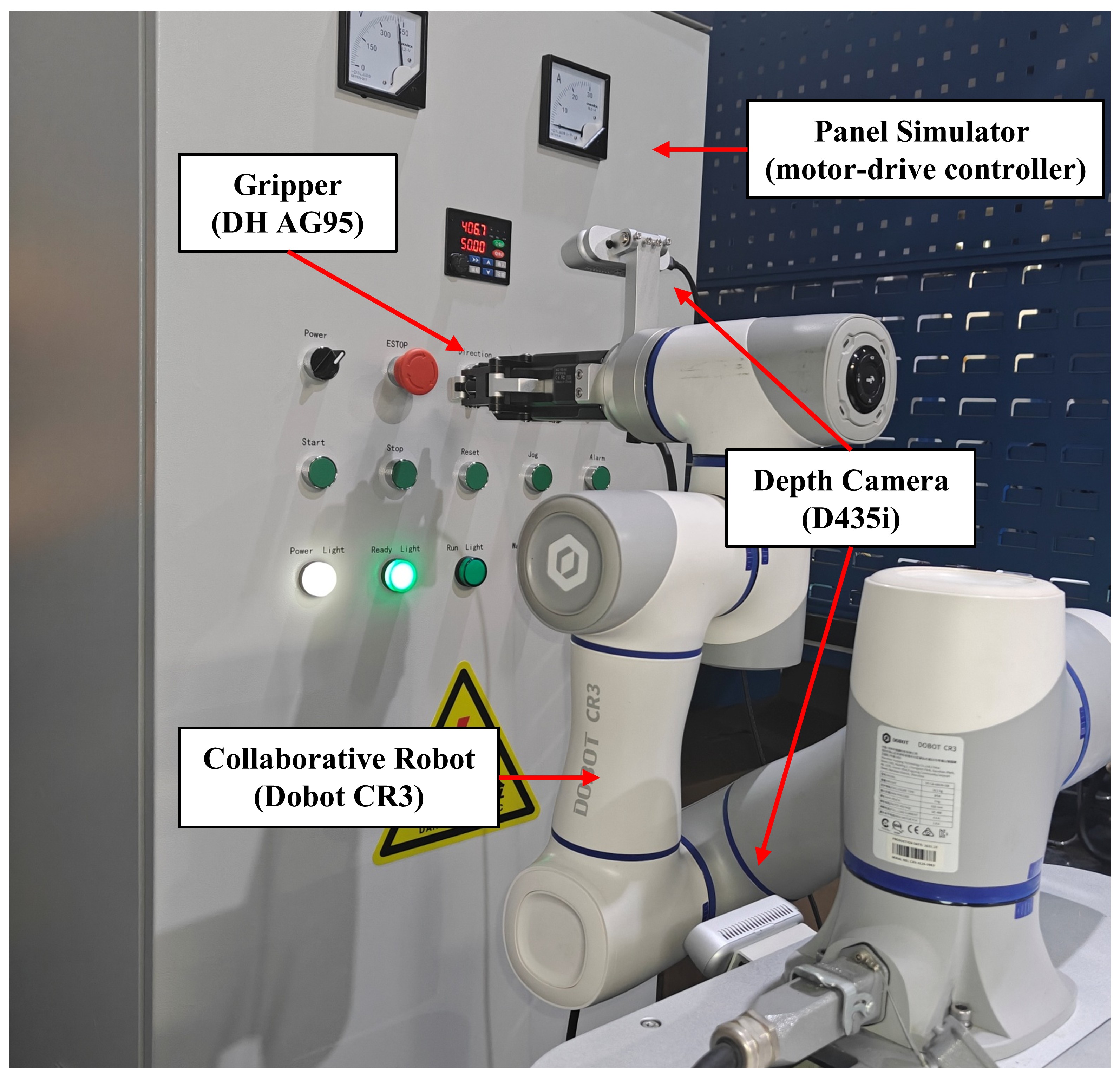}}
	\caption{Real-robot experimental platform for industrial panel operation.}
\label{pt_overview}
\end{figure}

As shown in Figure~\ref{pt_overview}, the robotic platform consists of a manipulator, a panel simulator of an industrial motor-drive controller, an Intel RealSense D435i depth camera, and upper-level planning/control nodes on ROS~2. No industrial load is attached. This configuration preserves the panel layout, control actuation, visual feedback, and device-state transitions needed to exercise the complete task pipeline, while excluding load-dependent process dynamics that are not modeled by the present task-level formulation. The perception and manipulation measurements are collected from five prescribed camera viewpoints, while the long-horizon evaluation comprises 25 executions from the nominal operating viewpoint. Across the five-view measurement study, the panel, camera calibration, and illumination are held fixed apart from the prescribed viewpoint change. The long-horizon trials use a fixed viewpoint and calibration; wear and cross-device hardware variation are not varied.

\begin{figure*}[!t]\centering
	\makebox[\linewidth][c]{\includegraphics[width=15cm]{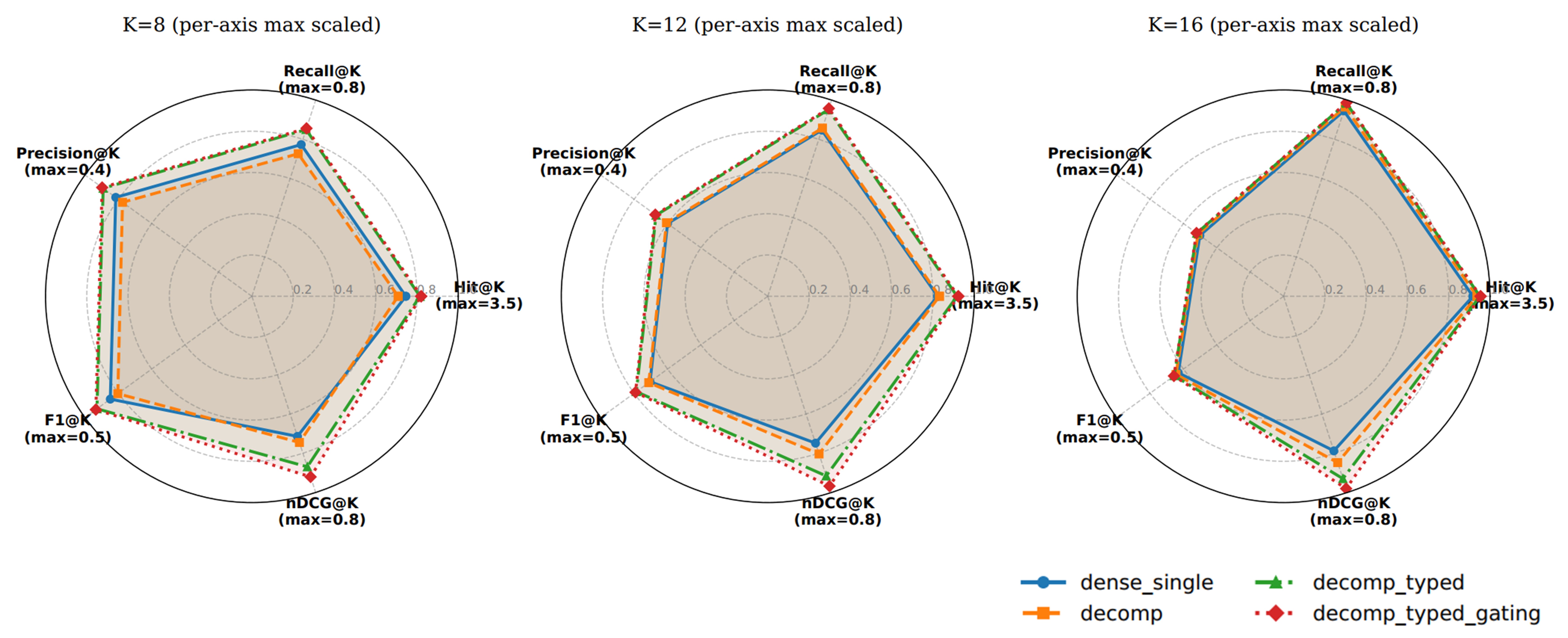}}
	\caption{Retrieval performance under cumulative retrieval ablations defined relative to Full \emph{MaCoPlanner}.}
\label{radar_overview}
\end{figure*}

\begin{figure}[!t]\centering
	\makebox[\linewidth][c]{\includegraphics[width=\columnwidth]{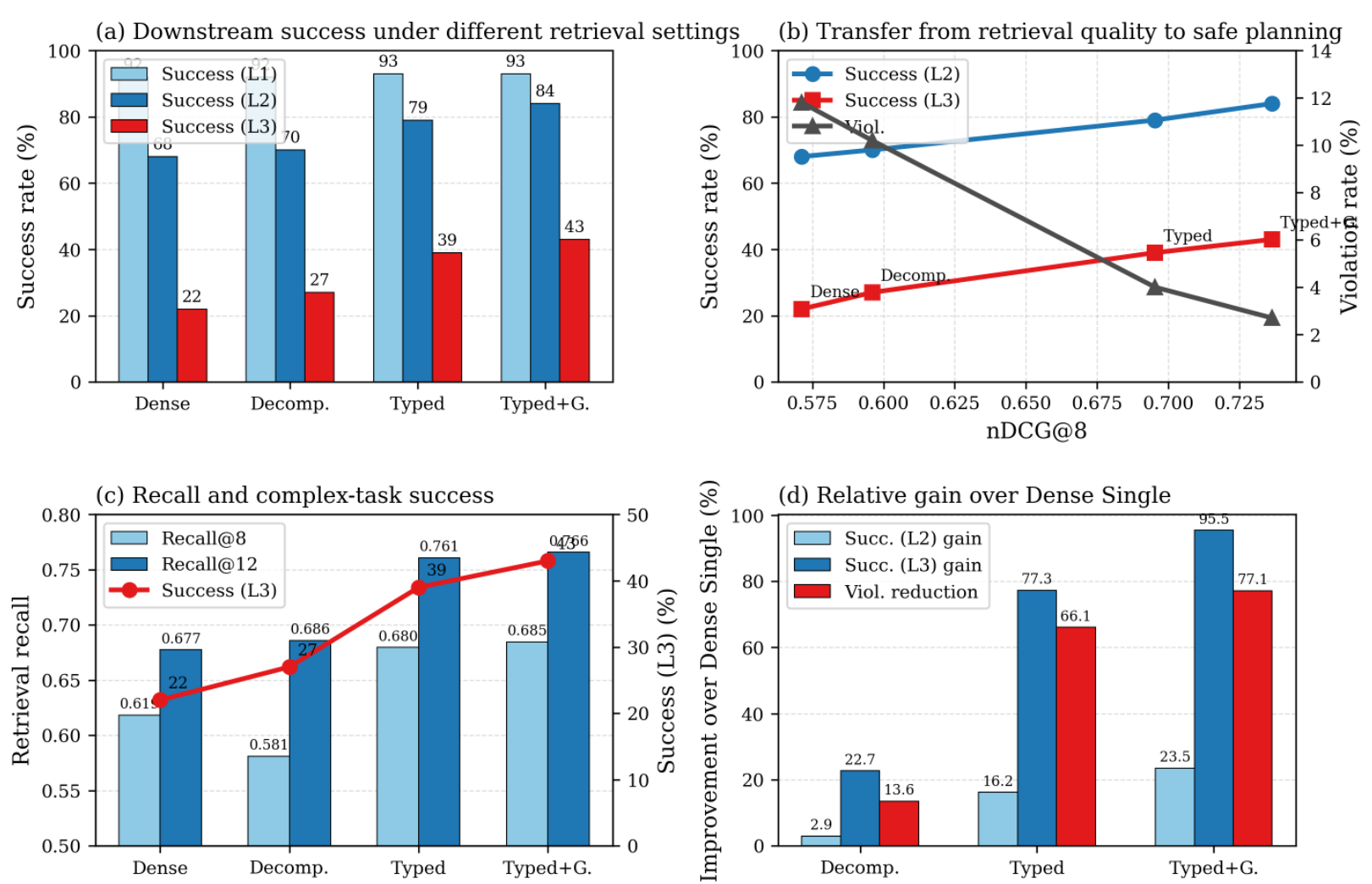}}
	\caption{Transfer from retrieval quality to downstream task-planning performance under proactive safety verification. The compact labels Dense, Decomp., Typed, and Typed+G. denote \emph{w/o Decomposition, Typed Indexing, and Evidence Gating}, \emph{w/o Typed Indexing and Evidence Gating}, \emph{w/o Evidence Gating}, and Full \emph{MaCoPlanner}, respectively.}
\label{tu7_overview}
\end{figure}

\subsubsection{Evaluation Metrics}
\textbf{Compilation fidelity metrics.}

We evaluate the generated representation against independently annotated source manuals rather than infer its quality from downstream planning performance. Completeness is the count-derived proportion of gold source units that have an aligned compiled item, whereas semantic consistency evaluates whether the aligned items preserve the original operational meaning and required fields.

\textbf{Retrieval and task planning metrics.}
The evaluation considers retrieval quality and task-level planning performance, including task completion, safety outcomes, and refinement behavior.
Retrieval quality is evaluated using \metric{Hit@K}, \metric{Recall@K}, \metric{Precision@K}, \metric{F1@K}, and \metric{nDCG@K}. 
Task success is measured by whether the generated primitive sequence reaches the benchmark target under the executable symbolic action semantics. Safety violation is evaluated using an independently annotated procedural/state oracle rather than the internal LTL specifications and FSM state-transition constraints used for repair. In addition to task success and safety violation, refinement behavior is evaluated by the average number of refinement steps. Fail-closed outcomes are described separately because rejection has different operational meanings across the compared methods.

\textbf{Perception and execution metrics.}
For physical interaction evaluation, measurement error evaluates the accuracy of the estimated geometric and interaction parameters, while operation success rate measures whether the executed action achieves the desired device response.
Long-horizon physical evaluation reports inference time, execution time, completion rate, and safety-trigger counts to characterize execution efficiency, task completion, and safety intervention during repeated operations.

\subsection{Compilation-Fidelity Evaluation of the Manual-to-IR Transformation}\label{subsec:compilation_fidelity}

Before evaluating retrieval quality, we directly assess whether the manual-compilation stage preserves the operational knowledge contained in the source manuals. The audit spans all four manual sources used by the retrieval and planning benchmarks: the motor-drive controller simulator, generator control system, laser controller, and engine controller. The exact compiled-IR snapshot used in the downstream experiments is aligned with a 350-unit reference set independently annotated from these four sources. This audit set is distinct from the full compiled-corpus inventory reported in Table~\ref{tab:corpus_stats}. \ref{app:benchmark_protocol} reports the annotation, agreement, and adjudication procedure.

The annotated units are grouped into process skeletons $\mathcal{T}$, safety clauses $\mathcal{S}$, parameter specifications $\mathcal{P}$, and structural relations $\mathcal{G}$. For each type, \metric{Completeness} is calculated directly as Compiled Items divided by Gold Units, while \metric{Semantic Consistency} is the proportion of aligned items that preserve the source meaning and required fields. Overall completeness uses the pooled counts, and overall semantic consistency is the macro-average across the four IR types.

\begin{table}[t]

\centering
\caption{Compilation-fidelity evaluation against independently annotated source manuals.}

\label{tab:compilation_fidelity}
\scriptsize
\renewcommand{\arraystretch}{0.7}
\setlength{\tabcolsep}{3pt}
{
\adjustbox{max width=\columnwidth}{
\begin{tabular}{lcccc}
\toprule
\tableheaderrow
\textbf{IR Type} & 
\textbf{Gold Units} &
\textbf{Compiled Items} &
\textbf{\metric{Completeness} (\%)} & 
\textbf{\metric{Semantic Consistency} (\%)} \\
\midrule
Process skeleton $\mathcal{T}$ & 37 & 35 & 94.59 & 97.14 \\
Safety clauses $\mathcal{S}$ & 81 & 81 & 100.00 & 97.53 \\
Parameters $\mathcal{P}$ & 54 & 53 & 98.15 & 98.11 \\
Relations $\mathcal{G}$ & 179 & 171 & 95.53 & 96.49 \\
\tablesectionrow
\textbf{Overall} & \textbf{350} & \textbf{340} & \textbf{97.14} & \textbf{97.32} \\
\bottomrule
\end{tabular}
}}
\end{table}

Table~\ref{tab:compilation_fidelity} shows that most annotated operational units have a corresponding compiled item. The count-derived coverage is complete for safety clauses, while the unmatched units are concentrated in process skeletons and structural relations, where procedural boundaries and cross-control dependencies must often be reconstructed from dispersed passages. Among the aligned items, semantic consistency remains similar across the four IR types; parameters perform best, whereas relations are more susceptible to incomplete or imprecise linkage. Thus, the two parts of the audit identify different failure modes---missing source requirements and semantic errors in retained items---and neither is used as a proxy for downstream retrieval or planning performance.

\subsection{Retrieval Quality Evaluation for Task-Relevant Manual Evidence}
We first evaluate whether the retrieval design supplies the evidence required for a task and its current device state. Using \textit{e5-base-v2}\cite{wang2022text} as the shared dense encoder, we take Full \emph{MaCoPlanner} as the reference and compare three cumulative retrieval ablations on the same compiled knowledge base: \emph{w/o Evidence Gating}, \emph{w/o Typed Indexing and Evidence Gating}, and \emph{w/o Decomposition, Typed Indexing, and Evidence Gating}. These conditions correspond to the compact plot labels Typed, Decomp., and Dense, respectively; Typed+G. denotes the full configuration. The candidate corpus and evaluation tasks are held fixed, so the comparisons isolate query decomposition, type-specific indexing, and evidence gating. The four typed indices correspond to process knowledge, parameter specifications, safety clauses, and control relations. The deployed index applies L2 normalization uniformly to both query and manual/evidence embeddings before similarity-based ranking.

Figure~\ref{radar_overview} shows only a limited improvement when query decomposition is restored, whereas restoring typed indexing produces the larger and more consistent improvement in the coverage-oriented metrics. For a fixed current device state, the typed subqueries are generated by the schema-constrained decomposition described in Section~\ref{subsubsec:query_decomp_retrieval}, validated by deterministic post-processing, and issued to their corresponding indices. Restoring evidence gating changes recall and F1 only slightly but improves \metric{nDCG@8} and \metric{nDCG@12} more clearly. The divergence between coverage and ranking metrics indicates that gating mainly promotes applicable, task-critical evidence within the retrieved context rather than expanding corpus coverage. In the evaluated implementation, a shared-threshold heuristic combines relevance with state applicability to produce this ordering.

\subsection{Impact of Retrieval Quality on Downstream Task-Planning Performance}\label{subsec:retrieval_transfer}
Figure~\ref{tu7_overview} further shows that the retrieval improvement is effectively transferred to downstream task planning under proactive safety verification, and that this effect becomes stronger as task complexity increases. The limited variation on Level-1 tasks indicates that simple single-control operations are less sensitive to retrieval quality. By contrast, the much larger differences on Level-2 and Level-3 tasks show that planning under cross-step dependency, interlock conditions, and explicit safety constraints relies heavily on structured and well-ranked evidence. In particular, Full \emph{MaCoPlanner} achieves the best downstream performance, improving Level-2 and Level-3 success while sharply reducing the violation rate relative to \emph{w/o Decomposition, Typed Indexing, and Evidence Gating}. In this transfer experiment, the planner, prompt, action space, state input, and knowledge source are fixed, and only the retrieval configuration is changed. Here, success is determined by whether the generated plan reaches the target state under executable action semantics, while violation rate is measured by the procedural and state-dependent constraint checks in Section~\ref{subsec:proactive_verification}. Overall, the proposed retrieval design improves not only retrieval quality itself, but also the usability of manual knowledge for constrained planning: typed indexing improves evidence completeness and type correctness, while evidence gating improves the visibility of task-critical constraints, both of which are reflected in better executability and more reliable task-planning outcomes under proactive safety verification.

\begin{figure}[!t]\centering
	\makebox[\linewidth][c]{\includegraphics[width=\columnwidth]{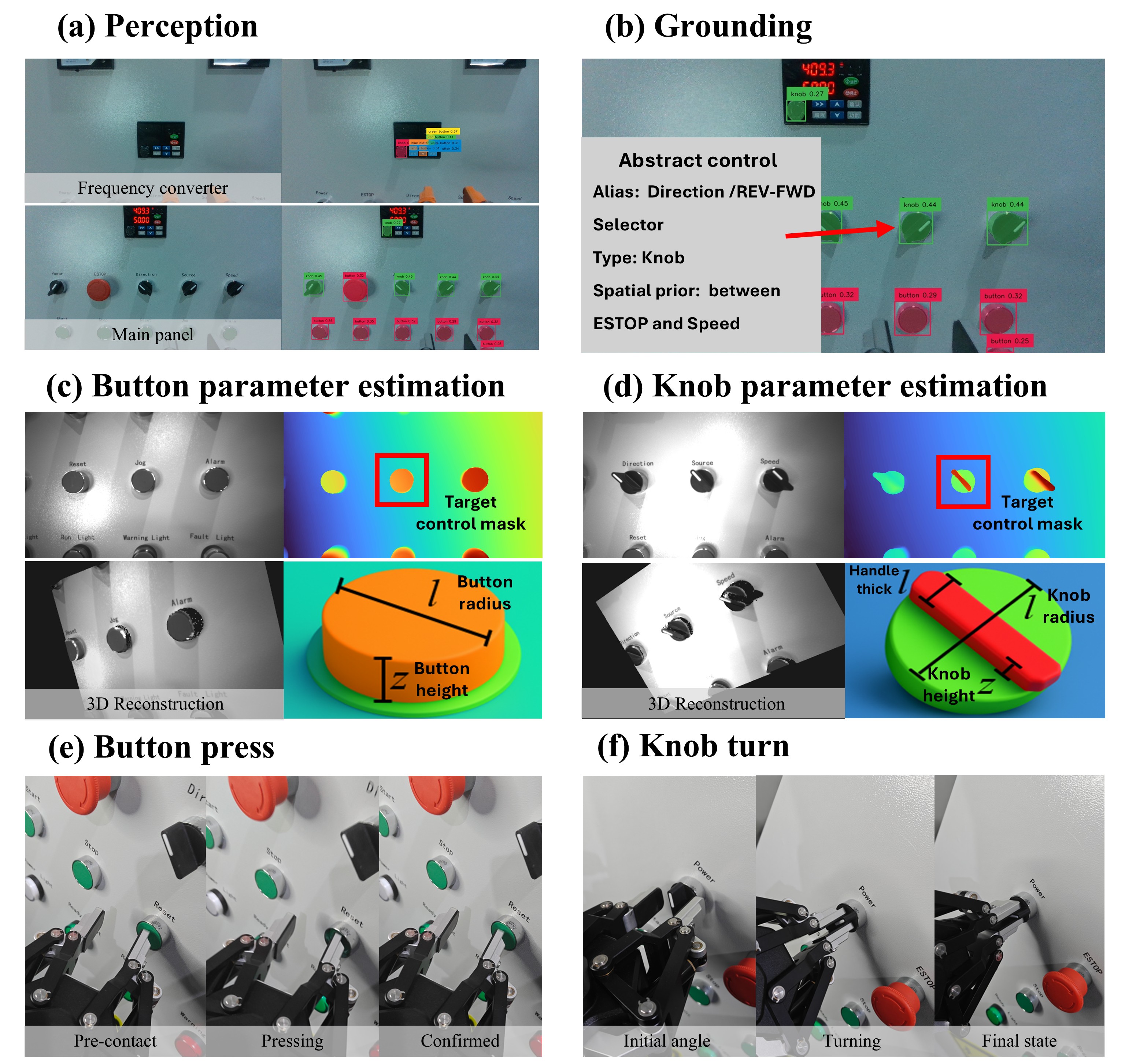}}
	\caption{Manual grounding, operation-oriented parameter estimation, and repeated manipulation examples.}
\label{tu8_overview}
\end{figure}

\subsection{Controlled Analysis of Knowledge Representation and System Components}\label{subsec:representation_scalability}

\begin{figure}[t]
\centering
\makebox[\linewidth][c]{\includegraphics[width=8.0cm]{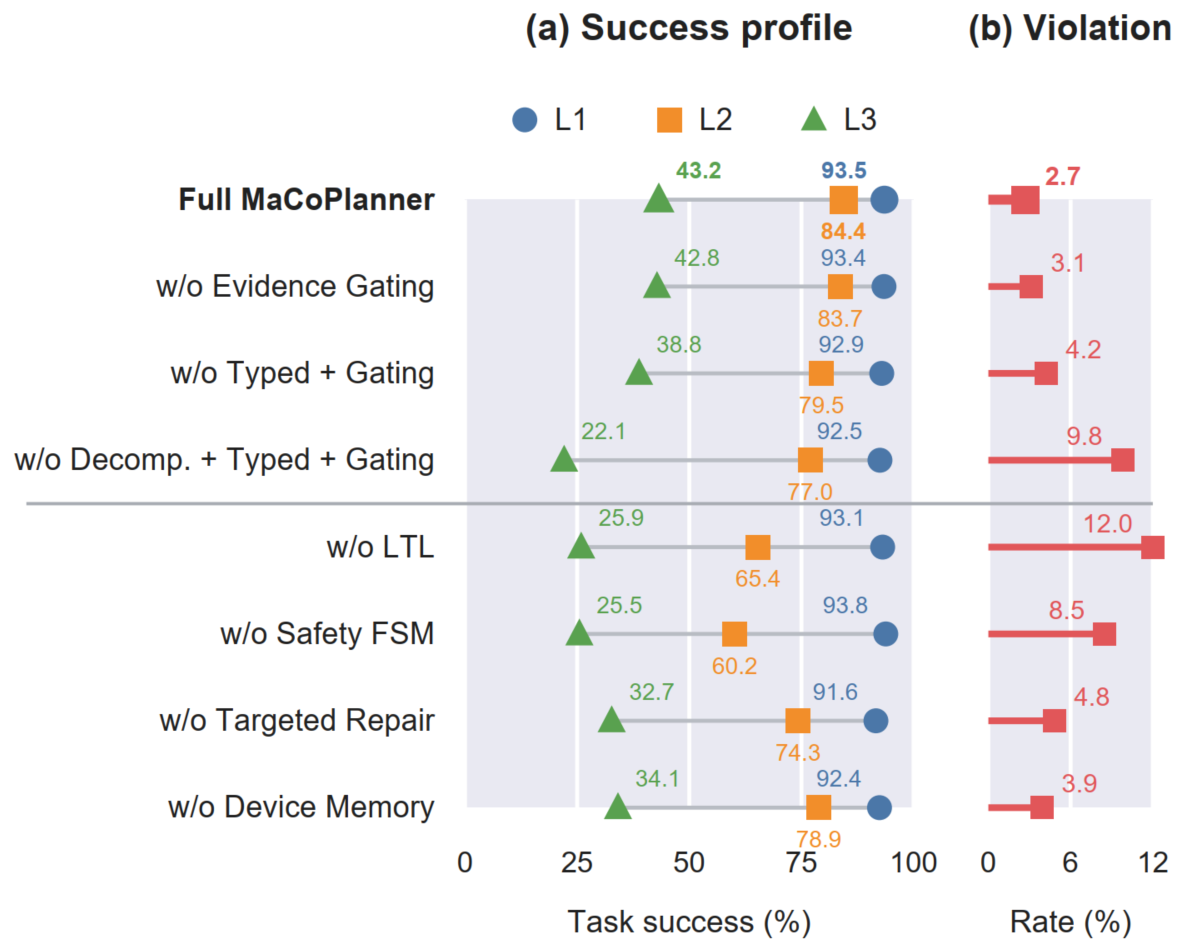}}
\caption{Controlled cumulative retrieval and system-component ablations in \emph{MaCoPlanner}. Colored markers and adjacent labels report exact success rates across the three task levels; the right panel reports the independent violation rate.}
\label{fig:controlled_analysis}
\end{figure}

\begin{table*}[t]

\centering
\caption{End-to-end comparison with representative task-planning baselines on the same planning benchmark.}
\label{tab:baseline_compare}
\tiny
\renewcommand{\arraystretch}{0.72}
\setlength{\tabcolsep}{3.5pt}
{
\adjustbox{max width=\textwidth}{
\begin{tabular}{
>{\arraybackslash}p{2.35cm}
>{\centering\arraybackslash}p{1.8cm}
>{\centering\arraybackslash}p{2.5cm}
>{\centering\arraybackslash}p{1.8cm}
>{\centering\arraybackslash}p{1.4cm}
c c c c c}
\toprule
\tableheaderrow
\textbf{Method} &
\textbf{Retrieval} &
\textbf{Formal verif.} &
\textbf{Repair} &
\textbf{Memory} &
\textbf{\metric{Succ.L1} (\%)} & 
\textbf{\metric{Succ.L2} (\%)} & 
\textbf{\metric{Succ.L3}(\%)} & 
\textbf{\metric{Viol.rate} (\%)} 
\\
\midrule
\emph{Raw-Manual Long-Context} 
& Raw manual 
& No
& No
& No
& 88.2 & 62.8 & 25.9 & 11.6  \\
\cmidrule(lr){1-9}
\emph{Prompt-Safety} 
& Text context 
& Nat.-lang. prompt 
& No
& No
& 93.1 & 70.5 & 33.8 & 8.3  \\
\cmidrule(lr){1-9}
\emph{ISR-LLM} \cite{zhou2024isr}
& Text context 
& Self-refine verifier 
& Iter. refine. 
& No
& 93.6 & 74.2 & 35.7 & 6.8  \\
\cmidrule(lr){1-9}
\emph{LLM}$^3$ \cite{wang2024llm}
& Task/motion fb. 
& Failure fb. 
& Motion-failure reas. 
& No
& 93.4 & 76.8 & 37.1 & 5.2 \\
\cmidrule(lr){1-9}
\emph{SafePlan} \cite{obi2025safeplan}
& Text context 
& Formal logic + safety 
& CoT refine.
& No
& 92.7 & 79.3 & 39.6 & 3.5  \\
\cmidrule(lr){1-9}
\tableaccentrow
\textbf{\emph{MaCoPlanner} (ours)} 
& Typed evidence 
& LTL + Safety FSM 
& Targeted repair 
& Device mem. 
& \textbf{93.5} & \textbf{84.4} & \textbf{43.2} & \textbf{2.7}  \\
\bottomrule
\end{tabular}
}}
\end{table*}

Figure~\ref{fig:controlled_analysis} expresses every controlled variant as a \emph{w/o} ablation of Full \emph{MaCoPlanner}. The retrieval rows use the same controlled factors as the transfer analysis in Figure~\ref{tu7_overview} and provide the representation-side reference for the component ablations. Their complexity-dependent pattern is consistent with the figure: Level-1 is comparatively insensitive, whereas removing typed organization and constraint-aware ranking produces progressively larger losses on Levels 2 and 3. Full \emph{MaCoPlanner} consequently achieves the strongest complex-task performance and the lowest independent violation rate.

The component ablations show the same complexity-dependent pattern. Removing LTL or the Safety FSM has little effect on simple Level-1 tasks but causes the largest degradation on Levels 2 and 3, indicating that temporal ordering and state-transition constraints become important only when plans contain coupled steps. Removing targeted repair or device memory also affects the more complex levels more strongly than Level-1, showing that constraint detection alone is insufficient without a mechanism for correction and an up-to-date representation of device state. The slight Level-1 increase without the FSM is therefore a local non-monotonic result rather than evidence that the FSM is unnecessary.

\subsection{End-to-End Task Planning and Task-Level Safety Evaluation}\label{subsec:end_to_end_eval}
After the controlled analyses, we compare the complete system with \emph{Raw-Manual}, \emph{Prompt-Safety}, \emph{ISR-LLM}, \emph{LLM}$^3$, and \emph{SafePlan}. Each baseline retains its native planning and safety mechanism, including iterative refinement only where it is part of the original method; \emph{MaCoPlanner}'s verifier is not imposed on the other methods. For each method, the evidence form, context budget, safety information, applicable refinement budget and stopping rule, and observed refinement usage are recorded in \ref{app:reproducibility}.

Table~\ref{tab:baseline_compare} reveals little separation among the advanced methods on Level-1, where \emph{MaCoPlanner} does not obtain the highest result. The performance gap emerges on Levels 2 and 3, where \emph{MaCoPlanner} achieves the strongest results among the comparable natural-language planning methods. Its widening advantage over \emph{Raw-Manual} indicates that direct use of unstructured manuals becomes increasingly inadequate as cross-step dependencies and safety constraints accumulate. Because the methods differ in evidence representation, verification, repair, and memory, their end-to-end differences cannot isolate individual component effects; those effects are examined in Figure~\ref{fig:controlled_analysis}.

\begin{figure}[t]
\centering
\makebox[\linewidth][c]{\includegraphics[width=8.0cm]{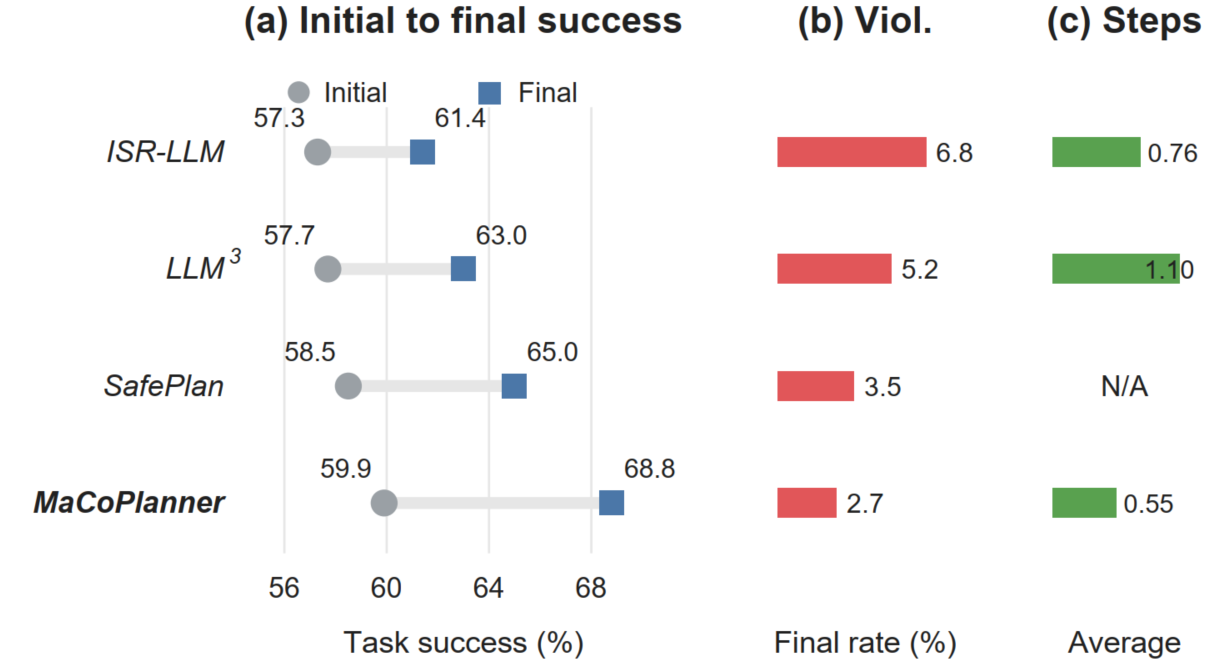}}
\caption{Initial and post-refinement outcomes for methods using post-generation verification or refinement. The dumbbell plot shows the change in task success; the adjacent panels report final violation rates and average refinement steps.}
\label{fig:first_final_outcomes}
\end{figure}

\emph{MaCoPlanner} nevertheless declines markedly on Level-3, showing that verification and repair cannot remove the underlying difficulty of generating valid plans under densely coupled prerequisites, interlocks, and evolving device states. Successful completion still depends on generating a valid alternative within the refinement budget. Figure~\ref{fig:first_final_outcomes} therefore compares initial task success with the final outcomes produced by post-generation safety checking or refinement. Because unresolved cases are handled differently by the baseline methods, rejection is discussed separately rather than treated as an additional cross-method metric.

\begin{figure}[!t]\centering
	\makebox[\linewidth][c]{\includegraphics[width=\columnwidth]{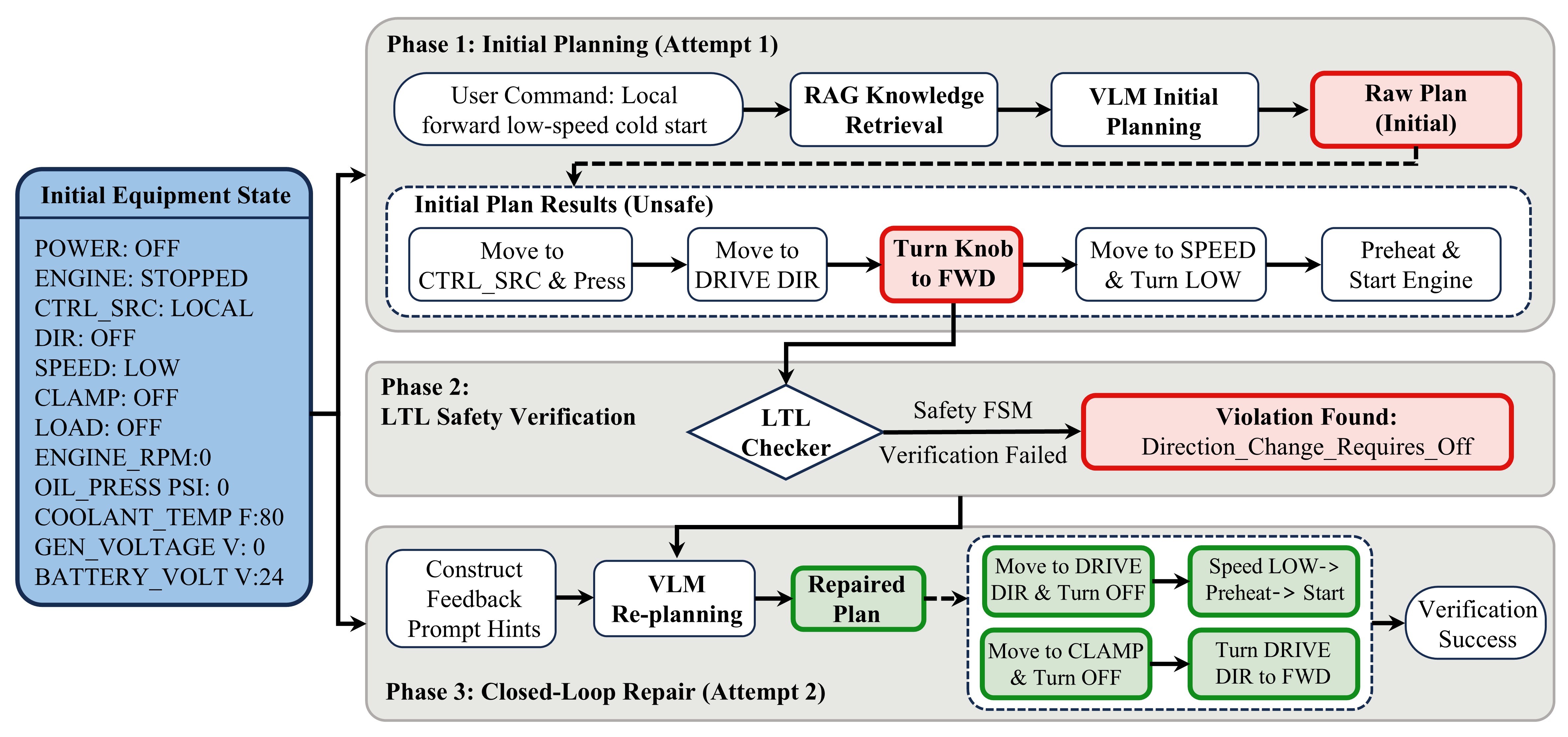}}
	\caption{Representative case showing the positions and roles of key modules in the end-to-end task-planning and proactive-verification pipeline.}
\label{tu12_overview}
\end{figure}

\begin{figure}[t]
\centering
\makebox[\linewidth][c]{\includegraphics[width=6.5cm]{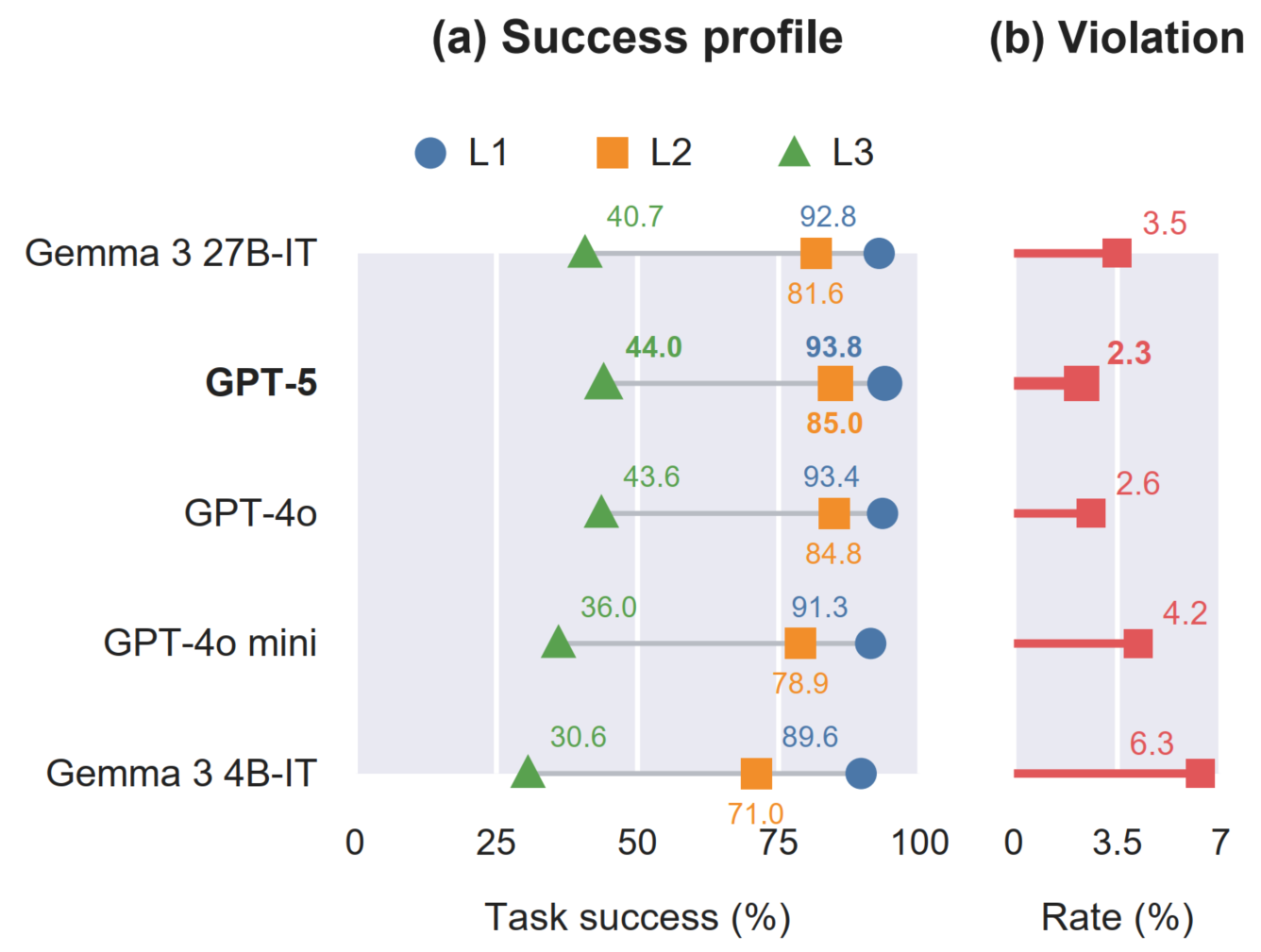}}
\caption{Backbone sensitivity on the 1/4 benchmark subset, including two pre-specified lower-capacity planners. Colored markers and adjacent labels report exact success rates across the three task levels; the right panel reports the violation rate.}
\label{fig:planner_backbone_results}
\end{figure}

Figure~\ref{fig:planner_backbone_results} shows that GPT-5 gives the highest point estimates, while GPT-4o remains close and Gemma~3 27B-IT exhibits a modest but increasingly visible deficit as task complexity rises. The two lower-capacity planners preserve much of the Level-1 performance but show larger losses on Levels 2 and 3 together with higher violation rates, indicating that the compiled representation and verification scaffold reduce, but do not remove, dependence on planner capacity. The comparison therefore supports robustness across the evaluated backbones without implying backbone independence beyond this subset. Figure~\ref{tu12_overview} complements these aggregate results with a representative trace through initial planning, pre-execution verification, violation localization, feedback construction, and targeted repair.

Figure~\ref{fig:first_final_outcomes} shows a consistent improvement from the initial to the final output after each method's native post-generation safety procedure. \emph{MaCoPlanner} achieves the strongest final success and the lowest final violation rate while using fewer refinement steps on average than the two explicit iterative baselines, \emph{ISR-LLM} and \emph{LLM}$^3$. This combination indicates that localized LTL/FSM feedback supports targeted correction rather than repeated unconstrained regeneration. \emph{SafePlan} is retained because its safety reasoning changes the final output, but its refinement count is marked N/A because it does not expose a comparable iterative plan-repair loop. The increase in \emph{MaCoPlanner}'s success remains smaller than the number of initially unresolved cases because verification cannot convert every failed candidate into a valid plan within the refinement budget.

Rejection is not reported as a separate cross-method metric because it has different operational meanings across the baselines. \emph{ISR-LLM} accepts the final generated action sequence when its iteration limit is reached \cite{zhou2024isr}, whereas \emph{LLM}$^3$ returns an unsuccessful planning outcome if no feasible task-and-motion solution is found within the prescribed attempts \cite{wang2024llm}. \emph{SafePlan} applies its native safety reasoning but does not implement fail-closed rejection after a bounded iterative repair loop. \emph{MaCoPlanner} instead applies a fail-closed rule to the generated plan: if no candidate satisfies both the LTL and Safety-FSM conditions within the refinement budget, no physical action is issued. Under this policy, 26.3\% of the 357 runs entering the planning loop were rejected. Thus, the post-refinement success gain should be interpreted together with the explicit suppression of unresolved plans, rather than as evidence that every detected failure can be repaired.

Figure~\ref{tu12_overview} illustrates how the aggregate repair behavior arises in an individual task. The initial plan attempts a direction change without completing the required OFF-state transitions, and the verifier localizes the corresponding constraint violation before execution. The resulting feedback leads the planner to insert the missing OFF operations and reorder the affected steps, after which the repaired plan satisfies both verification layers. The example therefore shows that repair is driven by a localized procedural or state-transition violation rather than by restarting generation without diagnostic feedback.

\subsection{Formal-Specification Fidelity and Planning Analysis}
The full source-to-specification audit in Table~\ref{tab:formal_audit} assesses the semantic fidelity of the LTL rules and FSM transitions against their source manuals. Semantically correct encodings are distinguished from approximations, local atomic-proposition or syntax errors, and LTL--FSM mismatches because each failure mode affects verification differently. The corresponding plan-level outcomes after verification and repair are reported in Figure~\ref{fig:first_final_outcomes}.

The PDDL experiment tests whether planning failures can be attributed to unreachable symbolic goals. Given the exact initial state, oracle symbolic goal, and complete action model, the planner returns a non-empty plan for every instance in the 50-task subset, establishing reachability within the encoded model. Because this diagnostic bypasses natural-language goal interpretation, manual-evidence retrieval, and construction of the symbolic planning problem, it is not treated as an end-to-end baseline. Its configuration is reported in \ref{app:reproducibility}. Level-2 and Level-3 plans have comparable lengths, whereas state-verification actions constitute 42\% of the Level-3 actions and are absent from the simpler levels. Complex tasks are therefore distinguished more by state confirmation and constraint coupling than by sequence length alone.

The remaining unsuccessful cases are examined together with grounding and execution failures in the failure-mode analysis below, allowing errors arising from evidence supply, plan verification, and physical interaction to be compared within a common pipeline-level taxonomy.

\subsection{Perception and Manipulation Validation}
The physical interface is evaluated separately from high-level task planning because a symbolically valid action can still fail if the target control is mislocalized or the interaction parameters are inaccurate. As illustrated in Figure~\ref{tu8_overview}, the grounding module maps abstract controls to panel instances and estimates the parameters used to instantiate pressing and rotation primitives. Each control category is measured from five prescribed camera viewpoints while the panel, depth-camera calibration, and illumination are held fixed. Table~\ref{tab:control_results} reports mean $\pm$ standard deviation across these five viewpoints for the applicable geometric or interaction deviations, together with the operation success rate pooled over repeated manipulation trials conducted at the same viewpoints. The number of views is therefore not the denominator of the success rate. Each repeated operation begins from the prescribed panel state and robot pre-contact pose, and success requires the expected observed state change after execution.

Table~\ref{tab:control_results} shows that ordinary button and knob operations combine small geometric deviations with high repeated-operation success, supporting the basic feasibility of manual-guided grounding. More importantly, localization error and operation success are not strictly monotonic across control types: small controls can have a lower absolute center deviation but slightly lower execution success than larger controls. Contact geometry, control shape, and interaction tolerance therefore affect execution in addition to localization accuracy. Because pressing and rotation involve different error dimensions, deviations are analyzed within each control type rather than pooled. The panel-mounted emergency-stop control is the clearest outlier, with the lowest operation success at 80.0\%. This experiment concerns manipulation of that panel control, not the robot or cell's independent protective emergency-stop function; the observed reliability is insufficient for unattended reliance on the panel control and remains a limitation of the physical execution layer.

\begin{table}[t]

\centering
\caption{Control-specific geometric/interaction deviations measured across five camera viewpoints and repeated manipulation success pooled over trials at the same viewpoints.}
\label{tab:control_results}
\scriptsize
\renewcommand{\arraystretch}{0.7}
\setlength{\tabcolsep}{3pt}
{
\adjustbox{max width=\columnwidth}{
\begin{tabular}{lccccccc}
\toprule
\tableheaderrow
\textbf{Control type} & \textbf{Views} & \textbf{Center dev. (mm)} & \textbf{Press dev. (mm)} & \textbf{Thickness dev. (mm)} & \textbf{Depth dev. (mm)} & \textbf{Rotation dev. (deg)} & \textbf{Succ. (\%)} \\
\midrule
Buttons & 5 & $0.80\pm0.04$ & $2.80\pm0.14$ & -- & -- & -- & 96.0 \\
Small buttons & 5 & $0.30\pm0.02$ & $1.00\pm0.05$ & -- & -- & -- & 93.0 \\
Knobs & 5 & $0.70\pm0.04$ & -- & $1.50\pm0.08$ & $2.60\pm0.13$ & $3.50\pm0.18$ & 95.0 \\
Emergency stop & 5 & $1.20\pm0.06$ & $3.50\pm0.18$ & $2.70\pm0.14$ & $3.40\pm0.17$ & -- & 80.0 \\
Small round knob & 5 & $0.40\pm0.02$ & -- & $1.10\pm0.06$ & $1.70\pm0.09$ & $2.20\pm0.11$ & 93.0 \\
\bottomrule
\end{tabular}
}}
\end{table}

\begin{table*}[t]

\centering
\caption{Planner/API latency, physical execution time, and task outcomes for the 25 long-horizon physical trials.}
\label{tab:real_chain_results}
\tiny
\renewcommand{\arraystretch}{0.72}
\setlength{\tabcolsep}{3.5pt}
{
\adjustbox{max width=\textwidth}{
\begin{tabular}{
>{\arraybackslash}p{3.2cm}
>{\centering\arraybackslash}p{1.8cm}
>{\centering\arraybackslash}p{1.8cm}
>{\centering\arraybackslash}p{1.7cm}
>{\centering\arraybackslash}p{1.7cm}
>{\centering\arraybackslash}p{1.7cm}
>{\centering\arraybackslash}p{2.2cm}}
\toprule
\tableheaderrow
\textbf{Task} &
\textbf{Planner/API (s)} &
\textbf{Execution (s)} &
\textbf{Completion} &
\textbf{1st attempt} &
\textbf{Avg. replans} &
\textbf{Avg. safety triggers} \\
\midrule

Turn on main power
& 3.5 & 4.1 & 5/5 & 5/5 & 0.0 & 0.0 \\
\cmidrule(lr){1-7}

Switch low speed under PANEL
& 4.2 & 3.9 & 5/5 & 4/5 & 0.2 & 0.0 \\
\cmidrule(lr){1-7}

Change the direction to REV
& 3.8 & 4.0 & 4/5 & 3/5 & 0.6 & 0.2 \\
\cmidrule(lr){1-7}

Perform a jog operation
& 3.7 & 2.5 & 5/5 & 4/5 & 0.2 & 0.2 \\
\cmidrule(lr){1-7}

Verify READY, then START
& 5.6 & 5.1 & 4/5 & 3/5 & 0.8 & 0.4 \\
\cmidrule(lr){1-7}

\tableaccentrow
\textbf{Aggregate}
& \textbf{4.2}
& \textbf{3.9}
& \textbf{23/25}
& \textbf{19/25}
& \textbf{0.36}
& \textbf{0.16} \\
\bottomrule
\end{tabular}
}}
\end{table*}

\begin{figure}[!t]\centering
\makebox[\linewidth][c]{\includegraphics[width=6cm]{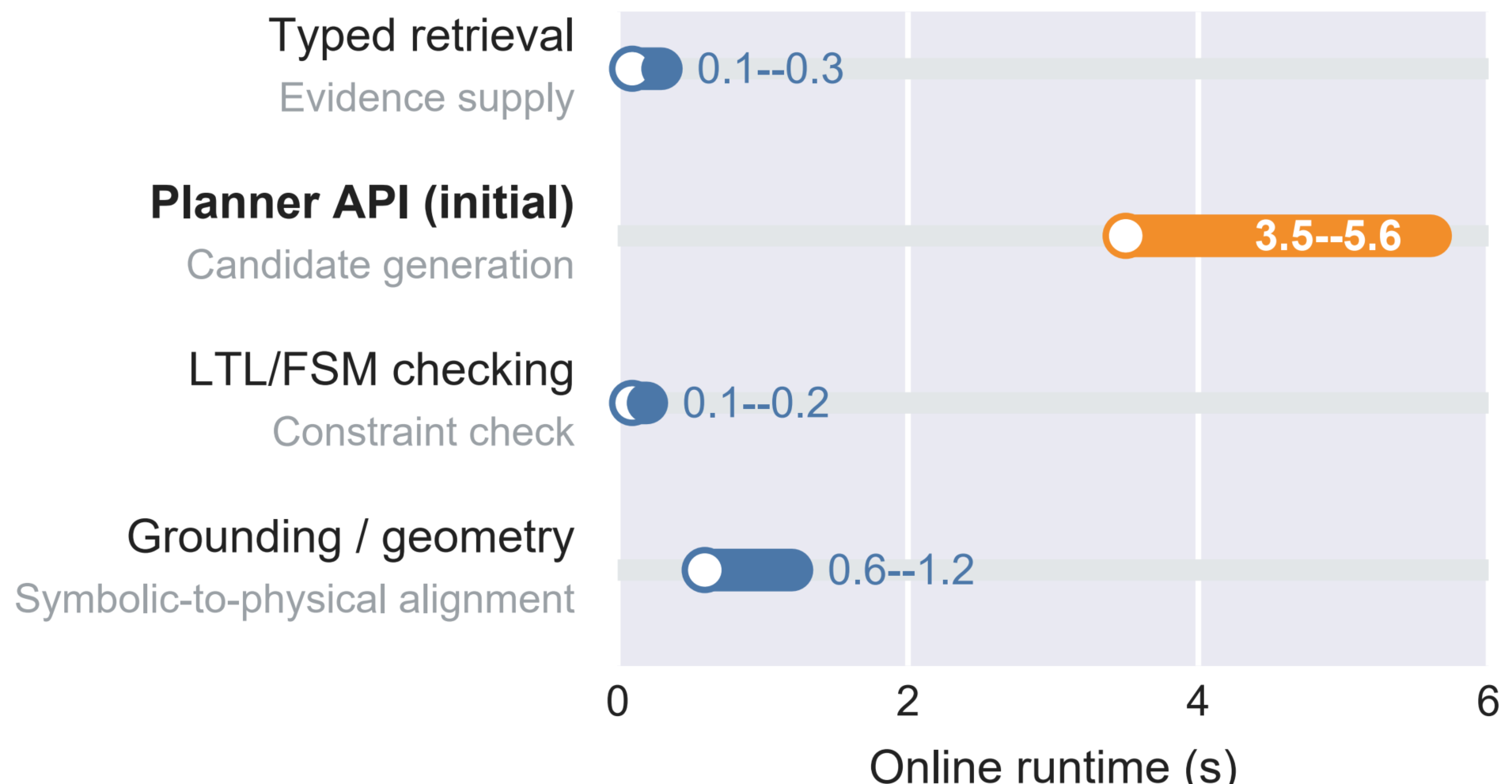}}
\caption{Module-level decomposition of online runtime for typed retrieval, the initial planner/API call, LTL/FSM checking, and grounding/geometry.}
\label{fig:deployment_cost}
\end{figure}

\subsection{Physical Experiments on Long-Horizon Tasks}
The long-horizon physical evaluation comprises five panel tasks, each repeated five times, for 25 executions in total. Before each run, the panel is restored to the prescribed task-specific initial state, the manipulator returns to the common pre-contact pose, and device memory is reinitialized from the current panel observation. The trials exercise planning, verification, grounding, execution, and state update under fixed panel, viewpoint, calibration, and illumination conditions; wear and cross-device variation are not varied.

Table~\ref{tab:real_chain_results} shows that most long-horizon trials complete successfully, with 23 of 25 reaching the target state. The lower first-attempt completion indicates that replanning recovers several trials rather than merely filtering failures. The simplest power-on task completes without intervention, whereas the direction-change and READY-to-START procedures concentrate the incomplete trials, repeated attempts, replanning, and safety triggers. This pattern is consistent with corrective planning being invoked primarily when execution depends on additional state confirmations and interlocks. Nevertheless, the two incomplete trials and the small physical sample prevent these results from being interpreted as a general reliability estimate.

\begin{figure}[!t]\centering
	\makebox[\linewidth][c]{\includegraphics[width=\columnwidth]{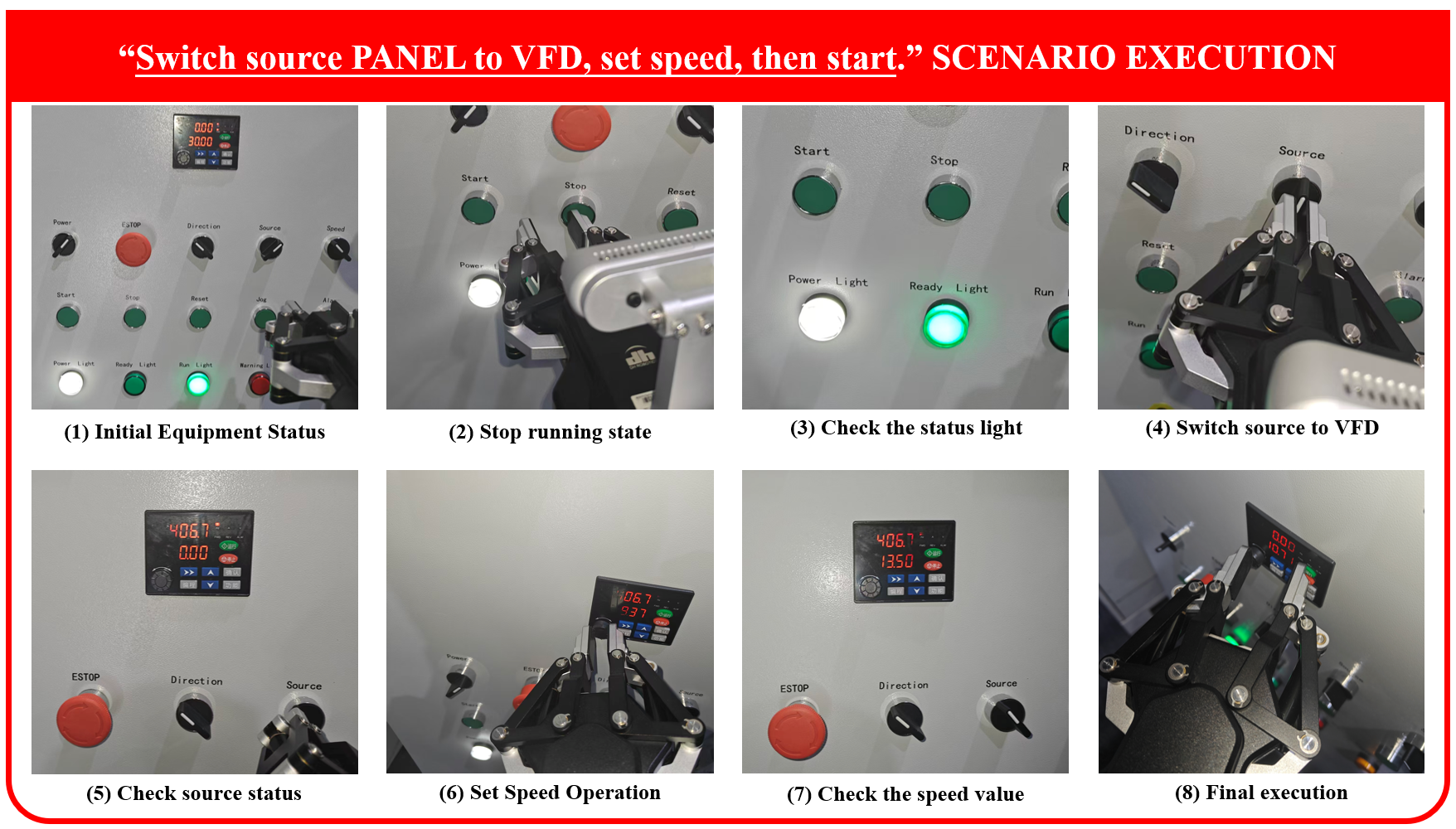}}
	\caption{Representative long-horizon execution sequence for switching the source to VFD, setting the speed, and starting the device.}
\label{tu11_overview}
\end{figure}

Figure~\ref{fig:deployment_cost} shows that the initial planner/API call dominates the reported computational cost, while typed retrieval and LTL-FSM checking contribute substantially less latency. Explicit task-level verification is therefore not the principal computational bottleneck in the measured pipeline. Grounding remains a non-negligible physical-interface cost, and repair generation is incurred only when violations are detected. The task-dependent physical execution times are reported separately in Table~\ref{tab:real_chain_results} and are not included in the module-level computational ranges. Because panel operation is a state-dependent supervisory procedure rather than a fixed-cycle repetitive production step, the measured latency is governed by device-specific response and confirmation requirements and does not constitute a general industrial cycle-time guarantee. Figure~\ref{tu11_overview} presents a representative long-horizon sequence in which the robot stops the running state, verifies the status indicator, changes the source to VFD, confirms the updated source state, sets and checks the speed, and then completes the final start operation.

\subsection{Failure Mode Analysis}
\begin{figure}[!t]\centering
	\makebox[\linewidth][c]{\includegraphics[width=\columnwidth]{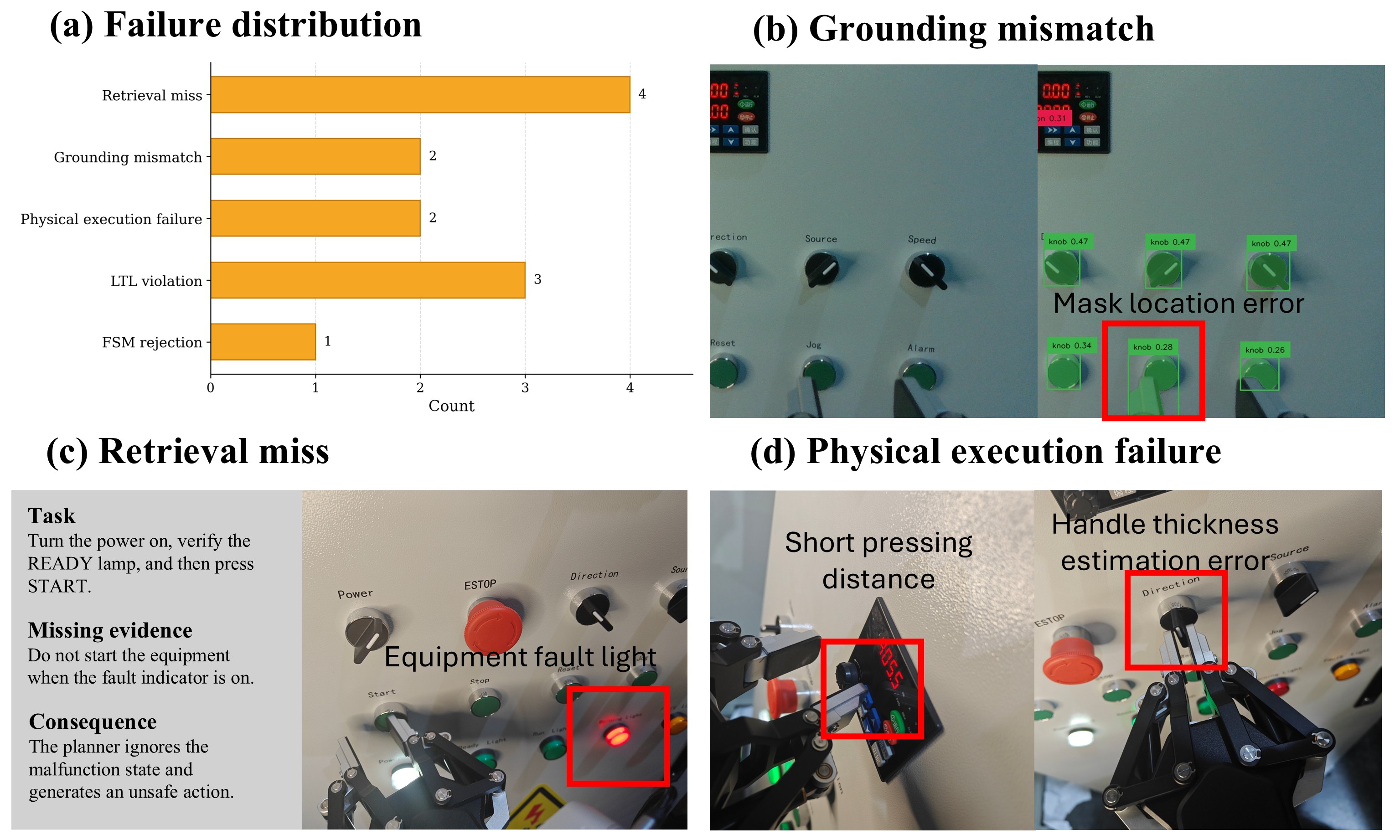}}
	\caption{Distribution of the 12 audited failed episodes and representative retrieval, grounding, and physical-execution failures.}
\label{tu10_overview}
\end{figure}

Each failed episode is assigned to the earliest identifiable breakdown represented in Figure~\ref{tu10_overview}: retrieval miss, LTL violation, FSM rejection, grounding mismatch, or physical execution failure. Across the 12 audited failed episodes, retrieval miss is the largest individual category. The remaining cases are evenly divided between plan-level verification outcomes (LTL violation or FSM rejection) and failures at the grounding/execution interface. This distribution shows that residual errors are not confined to candidate generation: incomplete evidence, formal constraint handling, and physical interaction all remain relevant to end-to-end performance.

Figure~\ref{tu10_overview}(b)--(d) illustrates the corresponding mechanisms. The grounding example shows how a mask-location error can associate an otherwise valid symbolic action with the wrong control. The retrieval example omits the evidence prohibiting startup when the fault indicator is active, leaving the planner without a task-critical state constraint. The physical examples show that insufficient pressing distance and handle-thickness estimation error can prevent the intended state transition even when the symbolic action is valid. These interface and execution failures lie outside the formal verifier's guarantee, which is conditional on a correct symbolic state and action-to-control correspondence. Algorithm~\ref{alg:macoplanner_online} rejects tasks for which no admissible symbolic plan is found before grounding; uncertainty introduced during grounding or execution instead requires stopping, reinitialization, or operator intervention beyond the symbolic verification layer.

\section{Conclusions}

This study presented \emph{MaCoPlanner}, a framework that combines typed knowledge compiled from equipment manuals with evidence-conditioned planning and proactive safety verification. Across 480 tasks from four panel domains, the framework improves complex task planning while maintaining a 2.7\% final violation rate under the independent safety oracle; plans that remain unresolved after repair are rejected before execution. Experiments on the no-load controller-panel simulator further demonstrate that verified symbolic plans can be grounded and executed under representative interaction conditions. The remaining limitations concern incomplete retrieval, symbolic-state drift, difficult Level-3 planning, and grounding or contact errors. Future work will strengthen these interfaces and evaluate the system under real loads and plant-level safety architectures.

\section*{CRediT authorship contribution statement}

Guipeng Xin: Writing--original draft, Validation, Software, Conceptualization. Jiahe Xu: Validation, Software. Mohammad Deghat: Writing--review \& editing, Supervision. Chenhui Wan: Software, Resources. Jie Liu: Resources, Investigation. Youmin Hu: Supervision, Project administration. Zhongxu Hu: Writing--review \& editing, Supervision, Project administration, Funding acquisition.

\section*{Declaration of competing interest}

The authors declare that they have no known competing financial interests or personal relationships that could have appeared to influence the work reported in this paper.

\section*{Acknowledgment}

\begingroup
\setlength{\emergencystretch}{2em}
This work was supported in part by the National Major Science and Technology Projects of China (No.~2025ZD1606204); the National Natural Science Foundation of China (No.~52575572 and 52575115); the Science and Technology Program of Hubei Province (No.~2024BEB025); the Joint Research Project of the Yangtze River Delta Science and Technology Innovation Community (No.~2024CSJGG1400); and the UNSW--HUST Global Research \& Impact Program (No.~5003100202).
\par
\endgroup

\section*{Data availability}

Code and data are publicly available at \url{https://github.com/XinGP/MaCoPlanner}.

\appendix
\section{Complete Prompts, Structured Outputs, and Post-Processing}

This appendix provides the prompts and structured output formats used by the language-driven modules, including manual compilation, query decomposition, task planning, violation-guided repair, memory update, and output post-processing. Table~\ref{tab:prompt_examples} consolidates the prompt-based stages in pipeline order. The compilation and rule-translation blocks correspond to Section~\ref{subsubsec:ir_construction}; query decomposition corresponds to Section~\ref{subsubsec:query_decomp_retrieval}; initial planning and violation-guided repair correspond to Sections~\ref{subsubsec:evidence_conditioned_planning} and~\ref{subsubsec:constraint_guided_repair}; and the memory-update and grounding blocks correspond to Section~\ref{subsec:grounding_memory}. Table~\ref{tab:output_validation} separately summarizes the deterministic validation and post-processing applied to their structured outputs.

\begin{table*}[t]

\centering
\caption{Consolidated prompt examples and instructions for the language-driven modules of \emph{MaCoPlanner}.}
\label{tab:prompt_examples}
\tiny
\renewcommand{\arraystretch}{0.72}
\setlength{\tabcolsep}{3.5pt}
{
\adjustbox{max width=\textwidth}{
\begin{minipage}{\textwidth}
\setlength{\parindent}{0pt}
\noindent\rule{\linewidth}{0.8pt}
\vspace{-0.3cm}

\textbf{Manual knowledge compilation}

You are a manual knowledge compiler for industrial equipment. Convert the provided manual chunk \texttt{\{manualChunk\}} into structured knowledge items for robotic panel operation. Classify each sentence into one of four types: procedure, safety, parameter, or relation. For a procedure item, extract \texttt{\{precondition, action, postcondition, retry\}}. For a safety item, extract \texttt{\{trigger, forbidden\_action, required\_condition\}}. For a parameter item, extract \texttt{\{name, unit, range, tolerance\}}. For a relation item, extract \texttt{\{source, relation, target\}}. Preserve the original terminology and attach a short evidence span from the manual. The required output format is: Type: \{...\}; Item: \{...\}; Evidence: \{...\}.

\vspace{0.08cm}
\noindent\rule{\linewidth}{0.5pt}

\textbf{Safety-rule translation}

You are a rule translator for industrial device operation manuals. Translate the given operation and safety text \texttt{\{manualRuleText\}} into formal logic constraints for plan verification. Focus on temporal order, forbidden transitions, mode-dependent interlocks, emergency handling, and operation prerequisites. Use the atomic proposition library \texttt{\{apLibrary\}} to write each rule as an LTL formula. If a rule cannot be formalized exactly, provide the closest conservative approximation and mark it explicitly. The required output format is: RuleID: \{...\}; Description: \{...\}; LTL: \{...\}; Approximate: \{Yes/No\}.

\vspace{0.08cm}
\noindent\rule{\linewidth}{0.5pt}

\textbf{Query decomposition and retrieval}

You are a retrieval planner for compiled manual knowledge. Given the user instruction \texttt{\{query\}} and the current device memory state \texttt{\{memoryState\}}, decompose the request into typed subqueries for procedure, safety, parameter, and relation retrieval. Each subquery should target one missing decision required for execution, and should prefer explicit control names, state variables, thresholds, and interlocks over generic paraphrases. Rank the subqueries by urgency and expected usefulness for planning. The required output format is: PROC: \{...\}; SAFE: \{...\}; PARAM: \{...\}; REL: \{...\}; Priority: \{...\}.

\vspace{0.08cm}
\noindent\rule{\linewidth}{0.5pt}

\textbf{High-level task planning}

You are a world-class robotic task planner that outputs ONLY executable robot action primitives for industrial panel operation. Given the user command \texttt{\{query\}}, device status \texttt{\{deviceState\}}, structured evidence \texttt{\{retrievedEvidence\}}, and the allowed action primitive library \texttt{\{primitiveLibrary\}}, generate a valid step-by-step task plan. Each step must satisfy the retrieved procedure constraints, safety rules, parameter limits, and current mode/interlock conditions. Prefer minimal and state-consistent operations, and avoid any action that changes direction, source, or speed under forbidden running states. The required output format is: StepX: \{Action primitive\}, \{Parameters\}.

\vspace{0.08cm}
\noindent\rule{\linewidth}{0.5pt}

\textbf{Violation-guided plan repair}

You are a violation-guided plan-repair module for industrial panel operation. Given the user goal \texttt{\{query\}}, current device state \texttt{\{deviceState\}}, retrieved evidence \texttt{\{retrievedEvidence\}}, previous candidate plan \texttt{\{candidatePlan\}}, localized LTL/FSM violation report \texttt{\{violationReport\}}, and allowed primitive library \texttt{\{primitiveLibrary\}}, revise only the steps needed to remove the reported violation. Preserve every valid prefix, insert missing prerequisites or verification actions when required, and do not introduce actions outside the primitive library. If the constraints cannot be satisfied, return \texttt{INFEASIBLE}. Otherwise, output only the revised sequence in the format: StepX: \{Action primitive\}, \{Parameters\}.

\vspace{0.08cm}
\noindent\rule{\linewidth}{0.5pt}

\textbf{Device-memory update}

You are a device-memory updater. Given the previous memory \texttt{\{memoryState\}}, executed primitive \texttt{\{executedAction\}}, post-action observation \texttt{\{observation\}}, and verification cue \texttt{\{verificationCue\}}, return a JSON object with the fields \texttt{mode}, \texttt{control\_states}, \texttt{parameters}, \texttt{last\_action}, \texttt{verification\_status}, \texttt{confidence}, and \texttt{timestamp}. Update only fields supported by the observation; retain unobserved fields but mark them stale when their age exceeds the staleness threshold. Do not overwrite a verified state with an unconfirmed inference. Confidence values must lie in $[0,1]$, and failed or missing verification must be recorded explicitly.

\vspace{0.08cm}
\noindent\rule{\linewidth}{0.5pt}

\textbf{Manual-guided grounding}

You are a location matcher for industrial panel grounding. You receive merged OCR annotations \texttt{\{ocrAnchors\}}, candidate control detections \texttt{\{detectedControls\}}, spatial priors \texttt{\{spatialRelations\}}, and target control labels \texttt{\{targetLabels\}}. OCR text anchors are soft hints only. Use semantic consistency, relative layout constraints, and 3D pose proximity to match each target label with the most likely physical control instance. For keypad-like layouts, same-row controls should have similar vertical coordinates and same-column controls should have similar horizontal coordinates in the robot base frame. Return the results only as compact mappings. The required output format is: \texttt{mapping: \{label: \{pose\_base\_m: \{x, y, z, frame\}\}\}}.

\vspace{0cm}
\noindent\rule{\linewidth}{0.8pt}
\end{minipage}
}}
\end{table*}

\begin{table*}[t]

\centering
\caption{Output validation and deterministic post-processing rules.}
\label{tab:output_validation}
\tiny
\renewcommand{\arraystretch}{0.72}
\setlength{\tabcolsep}{3.5pt}
{
\adjustbox{max width=\textwidth}{
\begin{tabular}{p{3.2cm}p{10.8cm}}
\toprule
\tableheaderrow
\textbf{Module} & \textbf{Validation / post-processing} \\
\midrule
Compilation & Parse the required type-specific schema; canonicalize units and enumerated values; require a non-empty source span; reject items with missing required fields, unresolved ambiguity, or unsupported values. \\
Query decomposition & Accept only \texttt{PROC}, \texttt{SAFE}, \texttt{PARAM}, and \texttt{REL}; retain at most one subquery per type; normalize absent types to an empty list; reject duplicated, unparseable, or out-of-schema fields. \\
Planner & Parse the ordered step sequence; require every action to belong to the primitive library; validate parameter names, types, ranges, and units; mark unknown, duplicated, or malformed actions invalid. \\
Violation feedback & Require each feedback item to resolve to a rule identifier, step index, and pre-action state; re-parse and re-verify every repaired plan before it can be accepted. \\
Memory update & Validate the memory schema and slot types; update only observation-supported slots; retain unobserved values with a stale flag; prevent low-confidence inferences from overwriting verified states. \\
Grounding & Require class/label compatibility and a valid base-frame pose; accept matches only when $S_{\mathrm{match}}\leq0.65$; use $\lambda=0.5$ for relation consistency and invoke the VLM only for unresolved semantic ambiguity. \\
\bottomrule
\end{tabular}
}}
\end{table*}

\section{Illustrative Examples of Safety Verification and Benchmark Construction}
\label{app:ltl_fsm_example}

This appendix provides concrete examples to further clarify two key parts of the proposed framework, namely the proactive safety verification mechanism and the benchmark-construction protocol. As described in Section~\ref{subsubsec:constraint_guided_repair} of the main text, proactive safety verification combines \emph{procedural constraint checking} and \emph{state-dependent legality checking}. In this appendix, the former is instantiated by representative LTL rules translated from manuals, while the latter is instantiated by a lightweight Safety FSM that tracks local engine operating modes. In addition, to make the benchmark design more transparent, a concrete IR-backed running example is further provided to show how procedure knowledge, safety constraints, control relations, and operation-critical parameter specifications are aligned into a unified benchmark item.

In the current implementation, each generated primitive sequence is first mapped into symbolic events, such as \texttt{pressed(ST}\\\texttt{ART\_BTN)}, \texttt{pressed(PANEL\_OFF)}. These symbolic events are then checked against the LTL rule set and further passed to the Safety FSM for local transition validation. Tables~\ref{tab:appendix_ltl_examples}--\ref{tab:appendix_violation_examples} summarize representative verification examples. Table~\ref{tab:benchmark_running_example} further provides a concrete IR-backed running example, showing how benchmark instances and reference targets are derived from the compiled manual knowledge.

Each action proposition in Table~\ref{tab:appendix_ltl_examples} is evaluated together with the symbolic state immediately before that action. The proposition \texttt{done(PREHEAT)} remains true after successful preheating.

\begin{table*}[t]

\centering
\caption{Representative LTL rules used in the safety-verification layer.}
\label{tab:appendix_ltl_examples}
\tiny
\renewcommand{\arraystretch}{0.72}
\setlength{\tabcolsep}{3.5pt}
{
\adjustbox{max width=\textwidth}{
\begin{tabular}{p{3.2cm} p{5.0cm} p{7.2cm}}
\toprule
\tableheaderrow
\textbf{Rule ID} & \textbf{Description} & \textbf{LTL formula} \\
\midrule
\makecell[tl]{\texttt{murphy\_contact\_}\\\texttt{required\_to\_run}}
& The engine can run only while the Murphy magnetic switch contact is latched closed. 
& \texttt{G ( is\_on(ENGINE) -> contact(MURPHY, CLOSED) )} \\

\makecell[tl]{\texttt{start\_requires\_}\\\texttt{coil\_deenergized}} 
& Cranking or running must not occur while the Murphy coil is energized. 
& \texttt{G ( (attempt\_start | is\_on(ENGINE)) -> !coil\_on(MURPHY) )} \\

\makecell[tl]{\texttt{preheat\_required\_for\_}\\\texttt{temperature\_below\_20C}} 
& Pre-heating is required when the operating temperature is below \(20^\circ\)C. 
& \texttt{G ( temp\_lt(20C) -> ( done(PREHEAT) \& light\_on(LED\_HEAT\_ORANGE) ) )} \\

\makecell[tl]{\texttt{drive\_direction\_}\\\texttt{mutual\_exclusion}}
& Forward and reverse drive commands must never be active at the same time. 
& \texttt{G ! ( is\_on(DRIVE\_FWD) \& is\_on(DRIVE\_REV) )} \\

\makecell[tl]{\texttt{switch\_on\_only\_}\\\texttt{if\_voltage\_ok}}
& The load may be switched on only when the generator voltage is within the permissible range. 
& \texttt{G ( switch\_on\_load -> ( voltage\_ok \& light\_on(LED\_AC\_OK\_GREEN) ) )} \\

\makecell[tl]{\texttt{switch\_off\_load\_first\_}\\\texttt{then\_press\_off} }
& When stopping the generator, the load should be switched off before pressing the OFF button. 
& \texttt{G ( pressed(PANEL\_OFF) -> is\_off(LOAD) )} \\
\bottomrule
\end{tabular}
}}
\end{table*}

\begin{table*}[t]

\centering
\caption{Representative Safety FSM used for local legality checking.}
\label{tab:appendix_fsm_examples}
\tiny
\renewcommand{\arraystretch}{0.72}
\setlength{\tabcolsep}{3.5pt}
{
\adjustbox{max width=\textwidth}{
\begin{tabular}{p{2.5cm} p{5.0cm} p{2.5cm} p{4.5cm}}
\toprule
\tableheaderrow
\textbf{Current state} & \textbf{Trigger / condition} & \textbf{Next state} & \textbf{Interpretation} \\
\midrule
\texttt{STOPPED} 
& \texttt{pressed(START\_BTN)} and no Murphy trip 
& \texttt{CRANKING} 
& Engine start has been requested and the system enters the cranking stage. \\

\texttt{CRANKING} 
& \texttt{ENGINE\_RPM > 400} 
& \texttt{RUNNING\_NO\_LOAD} 
& A successful start is confirmed by the engine-speed threshold while the load remains disconnected. \\

\texttt{CRANKING} 
& Murphy trip is active, or the main breaker is tripped 
& \texttt{FAULTED} 
& Unsafe start progression is interrupted and the system enters a fault state. \\

\texttt{RUNNING\_WITH\_LOAD} 
& \texttt{switch\_off(LOAD)} 
& \texttt{RUNNING\_NO\_LOAD} 
& The load is disconnected before the stop command is permitted. \\

\texttt{RUNNING\_NO\_LOAD} 
& \texttt{pressed(PANEL\_OFF)} and \texttt{is\_off(LOAD)} 
& \texttt{STOPPED} 
& The stop command is accepted only from the no-load running state. \\

\makecell[tl]{\texttt{RUNNING\_WITH\_LOAD}\\\texttt{RUNNING\_NO\_LOAD}}
& Murphy trip becomes active 
& \texttt{FAULTED} 
& A protection-triggered shutdown transfers the system into a faulted mode. \\

\texttt{FAULTED} 
& \texttt{pressed(RESET\_MURPHY)} 
& \texttt{STOPPED} 
& Manual reset clears the faulted condition and returns the system to a safe stopped state. \\
\bottomrule
\end{tabular}
}}
\end{table*}

\begin{table*}[t]

\centering
\caption{Representative implementation-level violation tags derived from rule and FSM checking.}
\label{tab:appendix_violation_examples}
\tiny
\renewcommand{\arraystretch}{0.72}
\setlength{\tabcolsep}{3.5pt}
{
\adjustbox{max width=\textwidth}{
\begin{tabular}{p{4.2cm} p{5.3cm} p{5.4cm}}
\toprule
\tableheaderrow
\textbf{Violation tag} & \textbf{Meaning} & \textbf{Typical corrective implication} \\
\midrule
\texttt{start\_requires\_murphy\_reset} 
& The plan attempts to start while the Murphy protection condition has not been reset. 
& Insert a reset step before the start command, or abort the start sequence. \\

\texttt{start\_requires\_sensor\_data} 
& Required oil-pressure or coolant-temperature sensing information is unavailable before start. 
& Add sensing/verification before allowing the start action. \\

\texttt{start\_requires\_preheat\_in\_cold} 
& The plan attempts to start in a cold condition without preheating. 
& Insert a preheat action and verify the corresponding indicator. \\

\texttt{direction\_change\_requires\_off} 
& The plan changes drive direction without first switching the drive to OFF. 
& Insert an intermediate \texttt{drive\_off} step before changing to FWD or REV. \\

\texttt{stop\_requires\_load\_off} 
& The plan presses the OFF button while the load is still ON. 
& Insert \texttt{load\_off} before the stop command. \\
\bottomrule
\end{tabular}
}}
\end{table*}

\begin{table*}[t]

\centering
\caption{IR-backed shutdown example illustrating source-to-task alignment and benchmark construction.}
\label{tab:benchmark_running_example}
\tiny
\renewcommand{\arraystretch}{0.72}
\setlength{\tabcolsep}{3.5pt}
{
\adjustbox{max width=\textwidth}{
\begin{tabular}{p{1.6cm} p{2.5cm} p{12cm}}
\toprule
\tableheaderrow
\textbf{Stage} & \textbf{IR source / field} & \textbf{Concrete IR knowledge entry} \\
\midrule

\multirow{2}{*}{\makecell[tl]{Task\\instance}}
& Task description
& \texttt{Stop the generator safely after unloading the load} \\

& State before
& \texttt{\{ENGINE\_RUNNING: true, LOAD: ON, PANEL\_ON: true, HIGH\_AMBIENT\_TEMP: false, HIGH\_PRIOR\_LOAD: false\}} \\
\midrule

\multirow{1}{*}{\makecell[tl]{Procedure\\skeleton}}
& \makecell[tl]{Process Framework\\(\texttt{TG-03})}
& \textbf{Name:} Stop the Generator (Unload $\rightarrow$ Cooldown $\rightarrow$ OFF) \newline
\textbf{Preconditions:} \texttt{Generator is running and needs to be shut down} \newline
\textbf{Steps:} \texttt{(1) Switch off load; (2) At high ambient temperature or high prior load, run without load for at least 5 minutes; (3) Press button 'off'; (4) Verify LED for 'on' is off.} \newline
\textbf{Postchecks:} \texttt{Generator is stopped; battery is not switched off prematurely; fuel valve closed only if needed.} \\
\midrule

\multirow{2}{*}{\makecell[tl]{Safety\\constraints}}
& \makecell[tl]{LTL constraint}
& \textbf{Description:} \texttt{Load must be turned off before shutting down the generator.} \newline
\textbf{Formula:}  \texttt{G ( shutdown(ENGINE) -> is\_off(LOAD) )} \\

& FSM constraint
& \textbf{State abstraction:} \texttt{\{RUNNING\_WITH\_LOAD, RUNNING\_NO\_LOAD, STOPPED\}} \newline
\textbf{Allowed transition:} \texttt{RUNNING\_WITH\_LOAD -> RUNNING\_NO\_LOAD -> STOPPED} \newline
\textbf{Forbidden transition:} \texttt{RUNNING\_WITH\_LOAD -> STOPPED} \newline
\textbf{Action mapping:} \texttt{switch\_off(load)} drives \texttt{RUNNING\_WITH\_LOAD -> RUNNING\_NO\_LOAD}; \texttt{press(OFF\_BUTTON)} drives \texttt{RUNNING\_NO\_LOAD -> STOPPED}. \\
\midrule

\multirow{1}{*}{\makecell[tl]{Control / relation\\graph}}
& Control--Step Graph
& \textbf{Edge 1:} \texttt{STEP\_SHUTDOWN\_1 -> CTRL\_LOAD\_SWITCH, rel=deactivates}, \texttt{detail="Switch off the load before shutdown"}. \newline
\textbf{Edge 2:} \texttt{STEP\_SHUTDOWN\_2 -> CTRL\_ENGINE\_SWITCH, rel=stops}, \texttt{detail="Press OFF only after LOAD=OFF"}. \newline
\textbf{View:} \texttt{shutdown\_procedure = \{key\_controls: [CTRL\_LOAD\_SWITCH, CTRL\_ENGINE\_SWITCH]\}}. \\
\midrule

\multirow{1}{*}{\makecell[tl]{Parameter\\specification}}
& Parameter Specs
& \textbf{Control part:} \texttt{Stop Generator Switch} (\texttt{Fischer Panda}, \texttt{10250088}) \newline
\textbf{Benchmark-membership status:} this example belongs to the generator-control benchmark domain represented by the power-management source manuals in Table~\ref{tab:exp_settings}; it is not an additional benchmark device. \newline
\textbf{Control family:} \texttt{button-like control} \newline
\textbf{Operation mode:} \texttt{press-to-stop} \newline
\textbf{Execution-critical parameters:} \texttt{\{target pose, approach direction, pressing depth/stroke, trigger threshold, dwell time\}} \newline
\textbf{Verification cue:} \texttt{LED turns off after actuation} \newline
\textbf{Function:} \texttt{Stops the generator, LED turns off} \\
\midrule

\multirow{2}{*}{\makecell[tl]{Benchmark\\target}}
& Gold evidence set
& \texttt{\{TG-03, shutdown\_requires\_load\_off, local shutdown FSM, G\_generator\_control\_step\_model, Stop Generator Switch\}} \\

& Reference plan target
& \texttt{switch\_off(load) -> [cooldown only if HIGH\_AMBIENT\_TEMP or HIGH\_PRIOR\_LOAD] -> press(OFF\_BUTTON) -> verify(LED\_ON = OFF)} \\
\bottomrule
\end{tabular}
}}
\end{table*}

\section{Independent Benchmark and Safety-Oracle Protocol}\label{app:benchmark_protocol}
This appendix records the independent annotation and audit procedures used for the retrieval benchmark, planning references, and external task-level safety oracle. The reference is constructed from the original manuals, task instruction, and initial device state, not copied from \emph{MaCoPlanner}'s compiled IR or internal verifier.

\begin{table*}[t]

\centering
\caption{Independent annotation, blinding, agreement, and adjudication protocol.}
\label{tab:annotation_protocol}
\tiny
\renewcommand{\arraystretch}{0.72}
\setlength{\tabcolsep}{3.5pt}
{
\adjustbox{max width=\textwidth}{
\begin{tabular}{p{3.1cm}p{11.8cm}}
\toprule
\tableheaderrow
\textbf{Item} & \textbf{Protocol} \\
\midrule
Annotator count / expertise & 3 / two industrial-automation researchers and one robotics researcher \\
Materials shown & Original manuals + task instruction + initial device state; \emph{MaCoPlanner} retrieval/IR output hidden \\
Task-required evidence & Complete source spans needed to determine the correct action/decision \\
State-applicable evidence & Subset valid under the specified device state \\
Independent safety oracle & Procedural/state constraints independently annotated from source manuals and task conditions \\
Blinding & Items were independently annotated in randomized order; annotators did not see \emph{MaCoPlanner} outputs or the other annotators' labels. \\
IAA metric(s) & Fleiss' $\kappa$ for categorical decisions and span-level F1 for source-evidence selection \\
Adjudication & Disagreements were resolved by consensus after source review; unresolved cases were decided by a senior domain expert. \\
Benchmark audit & Check evidence, initial state, conditional branches, reference plan, safety requirements, and control relations against source manuals \\
\bottomrule
\end{tabular}
}}
\end{table*}

\section{Reproducibility and Implementation Details}\label{app:reproducibility}
This appendix consolidates the experiment-critical implementation settings used for manual compilation, retrieval, planning, verification, grounding, baseline evaluation, and stateless API inference.

\begin{table*}[t]

\centering
\caption{Consolidated implementation and hyperparameter configuration.}
\label{tab:repro_settings}
\tiny
\renewcommand{\arraystretch}{0.72}
\setlength{\tabcolsep}{3.5pt}
{
\adjustbox{max width=\textwidth}{
\begin{tabular}{p{3.4cm}p{4.2cm}p{7.2cm}}
\toprule
\tableheaderrow
\textbf{Component} & \textbf{Setting} & \textbf{Value / implementation} \\
\midrule
Manual chunking & chunk size / overlap & 512 / 64 tokens \\
Compilation model & exact model snapshot/API & \texttt{gpt-4o} through the OpenAI API \\
Query decomposition & model / subqueries & \texttt{gpt-4o} / at most one per type \\
Dense retriever & encoder & e5-base-v2 \\
Embedding normalization & query + manual item & L2 normalization for both query and manual/evidence embeddings \\
Retrieval & $L$, $K$ & $L=5K$ (40/60/80 candidates); evaluation $K\in\{8,12,16\}$ \\
Evidence gating & $\alpha$, $\tau$, applicability logic & $\alpha=0.70$; $\tau=0.55$; hard state contradictions are vetoed; the remaining mode, control, state, and parameter checks use equal weights; shared $\tau_c=\tau$ \\
Planner API & exact model snapshots & GPT-4o: \texttt{gpt-4o}; GPT-5: \texttt{gpt-5}; Gemma 3: \texttt{google/gemma-3-27b-it} and \texttt{google/gemma-3-4b-it} \\
Planner API state & request isolation & Fresh context and no cross-task conversational memory \\
Refinement & $K_{\max}$ & 3 \\
Grounding score & $\lambda$ / matching threshold & $0.5$ / $0.65$; fixed before evaluation; relation consistency remains secondary to class/OCR and geometric filtering \\
Geometry & DBSCAN/plane/shape parameters & DBSCAN: $\epsilon_{\mathrm{db}}=0.015$ m, minimum samples $=10$; plane: RANSAC distance $=0.005$ m; shape: circularity $\geq0.75$, aspect ratio $\in[0.8,1.25]$ \\
Device memory & schema / stale-state handling & JSON slots \texttt{\{mode, control\_states, parameters, last\_action, verification\_status, confidence, timestamp\}} / staleness threshold: 5 s \\
Symbolic planner & search algorithm / limits & Fast Downward with \texttt{lama-first} / 60 s and 4 GB per task \\
\bottomrule
\end{tabular}
}}
\end{table*}

\noindent\textbf{Rationale and scope of the fixed settings.}
The 512-token chunk length follows the maximum input length of the deployed \textit{e5-base-v2} encoder, while the 64-token overlap (12.5\% of a chunk) retains clauses crossing adjacent chunk boundaries without duplicating most of the corpus. L2-normalized query and evidence embeddings follow the reference E5 inference procedure, making their dot product equivalent to cosine similarity~\cite{wang2022text,e5basev2modelcard}. At most one subquery is generated for each IR type because decomposition is constrained by the four-slot schema rather than selected by an unconstrained search. The reported $K\in\{8,12,16\}$ values are evaluation cutoffs spanning three compact evidence budgets, and $L=5K$ deliberately supplies a larger candidate pool for applicability gating before the final top-$K$ context is formed; these values are held fixed across retrieval variants.

For evidence gating, $\alpha=0.70$ intentionally prioritizes semantic relevance while retaining a 30\% state-applicability contribution. The shared $\tau=0.55$ cutoff is a conservative near-midrange filter used after the independent hard-conflict veto. The two values were fixed before evaluation and applied to every evidence type and retrieval variant; no test-task tuning or claim of universal optimality is made.

The repair cap $K_{\max}=3$ permits at most three verifier-guided repair calls after the initial candidate, consistent with bounded iterative-refinement designs~\cite{madaan2023selfrefine}, while limiting latency and enforcing fail-closed rejection when no candidate passes both checkers. It is a fixed system budget rather than a literature-standard optimum.

In grounding, $\lambda=0.5$ deliberately makes relation consistency a secondary correction to the pose term, and the $0.65$ matching-cost threshold retains plausible candidates after class/OCR and geometric filtering so that only unresolved ambiguity is delegated to the VLM. Both are fixed engineering settings used unchanged across the five camera viewpoints. DBSCAN and RANSAC define the clustering and robust-fitting procedures~\cite{ester1996dbscan,fischler1981ransac}, but their numerical thresholds are necessarily scale- and noise-dependent. Here, the 15-mm DBSCAN radius groups points at individual-control scale, the ten-point minimum suppresses isolated depth returns, and the 5-mm RANSAC distance defines plane inliers; the circularity and aspect-ratio limits encode the nominal round or near-square control shapes. The 5-s stale-state threshold permits one perception--action--confirmation cycle but prevents an unrefreshed state from being reused indefinitely. Finally, \texttt{lama-first} follows the established Fast Downward/LAMA first-solution configuration~\cite{helmert2006fastdownward,richter2010lama}; 60 s and 4 GB are uniform per-task resource caps. The 16,000-token context budget is matched across neural baselines for comparison, whereas the ISR-LLM and \emph{LLM}$^3$ iteration/call limits retain their cited native interfaces~\cite{zhou2024isr,wang2024llm} and are not tuned as \emph{MaCoPlanner} parameters.

\begin{table*}[t]

\centering
\caption{Structure of the per-trial physical, runtime, and per-task planning records.}
\label{tab:raw_data_placeholder}
\tiny
\renewcommand{\arraystretch}{0.72}
\setlength{\tabcolsep}{3.5pt}
{
\adjustbox{max width=\textwidth}{
\begin{tabular}{p{4.0cm}p{10.5cm}}
\toprule
\tableheaderrow
\textbf{Record} & \textbf{Fields} \\
\midrule
Per-control physical measurements & Trial identifier, control type, calibrated reference, predicted center/press/depth/thickness/rotation values, success outcome \\
Long-horizon runtime & Task/trial ID and timestamps for retrieval, planner, rollout, LTL/FSM, feedback construction, repair API, grounding, execution, state confirmation \\
Planning outcomes & Task ID, method, initial/final plan status, independent violation, refinements, and fail-closed outcome \\
\bottomrule
\end{tabular}
}}
\end{table*}

\begin{table*}[t]

\centering
\caption{Retrieval/planning benchmark composition and independent annotation statistics.}
\label{tab:benchmark_audit_stats}
\tiny
\renewcommand{\arraystretch}{0.72}
\setlength{\tabcolsep}{3.5pt}
{
\adjustbox{max width=\textwidth}{
\begin{tabular}{lcccccc}
\toprule
\tableheaderrow
\textbf{Benchmark / source} & \textbf{Tasks} & \textbf{Device/manual split} & \textbf{Single-/multi-evidence} & \textbf{Annotators} & \textbf{IAA} & \textbf{Adjudicated items} \\
\midrule
Retrieval benchmark & 400 & 100/100/100/100 & 144/256 & 3 & $\kappa=0.87$; span F1$=0.90$ & 47 \\
Planning benchmark L1/L2/L3 & 80/200/200 & 120/120/120/120 & n/a & 3 & $\kappa=0.91$ & 31 \\
Independent safety oracle & 480 task evaluations & same task split & n/a & 3 & $\kappa=0.93$ & 18 \\
\bottomrule
\end{tabular}
}}
\end{table*}

\begin{table*}[t]

\centering
\caption{Full-corpus source-manual and compiled-IR provenance statistics across the four benchmark domains. These counts are distinct from the independently annotated audit subset in Table~\ref{tab:compilation_fidelity}. All manuals use the same 512-token chunking window with a 64-token overlap.}
\label{tab:corpus_stats}
\tiny
\renewcommand{\arraystretch}{0.72}
\setlength{\tabcolsep}{3.5pt}
{
\adjustbox{max width=\textwidth}{
\begin{tabular}{lcccccccc}
\toprule
\tableheaderrow
\textbf{Device/manual source} & \textbf{Manuals} & \textbf{Pages} & \textbf{Chunks} & $|\mathcal{T}|$ & $|\mathcal{S}|$ & $|\mathcal{P}|$ & $|\mathcal{G}|$ & \textbf{Compiled total} \\
\midrule
Motor-drive controller simulator & 1 & 57 & 65 & 16 & 41 & 33 & 65 & 155 \\
Generator control system & 1 & 42 & 48 & 12 & 21 & 9 & 18 & 60 \\
Laser controller & 1 & 92 & 105 & 7 & 19 & 11 & 88 & 125 \\
Engine controller & 1 & 113 & 129 & 15 & 26 & 17 & 65 & 123 \\
\midrule
\tablesectionrow
\textbf{Overall} & \textbf{4} & \textbf{304} & \textbf{347} & \textbf{50} & \textbf{107} & \textbf{70} & \textbf{236} & \textbf{463} \\
\bottomrule
\end{tabular}
}}
\end{table*}

\begin{table*}[t]

\centering
\caption{Baseline information access, context, safety mechanism, and refinement configuration.}
\label{tab:baseline_budget_matrix}
\tiny
\renewcommand{\arraystretch}{0.72}
\setlength{\tabcolsep}{3.5pt}
{
\adjustbox{max width=\textwidth}{
\begin{tabular}{lccccccc}
\toprule
\tableheaderrow
\textbf{Method} & \textbf{Evidence form} & \textbf{Context budget} & \textbf{Safety information} & \textbf{Verifier / feedback} & \textbf{Max. refinements} & \textbf{Stop/reject rule} & \textbf{Same typed evidence?} \\
\midrule
\emph{Raw-Manual} & Raw manuals & 16,000 input tokens & source text & No & No & single-pass output & No \\
\emph{Prompt-Safety} & Text context & 16,000 input tokens & prompt rules & prompt only & No & single-pass output & No \\
\emph{ISR-LLM} & Text context & 16,000 input tokens & prompt-level safety rules & self-refine & 3 & return the last sequence at the iteration cap; no fail-closed rejection & No \\
\emph{LLM}$^3$ & PDDL domain/problem and text context & 16,000 input tokens & PDDL preconditions/effects and feasibility checks & failure feedback & 10 calls & stop at the call budget; no independent fail-closed rejection & No \\
\emph{SafePlan} & Text context & 16,000 input tokens & formal/natural safety information & CoT-guided safety reasoning & N/A & no fail-closed rejection after safety reasoning & No \\
\tableaccentrow
\textbf{\emph{MaCoPlanner}} & Typed compiled evidence & 16,000 input tokens & compiled task constraints & LTL+FSM feedback & 3 & reject if unverified & Yes \\
\bottomrule
\end{tabular}
}}
\end{table*}

\begin{table*}[t]

\centering
\caption{Full LTL/FSM rule audit against source manuals.}
\label{tab:formal_audit}
\tiny
\renewcommand{\arraystretch}{0.72}
\setlength{\tabcolsep}{3.5pt}
{
\adjustbox{max width=\textwidth}{
\begin{tabular}{lcccccc}
\toprule
\tableheaderrow
\textbf{Rule family} & \textbf{Total} & \textbf{Approximate} & \textbf{Semantically correct} & \textbf{AP/syntax issue} & \textbf{Cross LTL--FSM mismatch} & \textbf{Corrected} \\
\midrule
LTL rules & 81 & 6 & 75 & 3 & 3 & 6 \\
FSM transitions / guards & 48 & n/a & 45 & 1 & 2 & 3 \\
\bottomrule
\end{tabular}
}}
\end{table*}

\bibliographystyle{elsarticle-num}
\bibliography{reference_stiima}

\end{document}